\documentclass[times,twocolumn,final]{article}%{elsarticle}
\usepackage{amssymb}
\usepackage{latexsym}
\usepackage{graphicx}
\usepackage[style=numeric,backend=biber,bibencoding=utf-8]{biblatex}
\usepackage{float}

\usepackage{url}
\usepackage{xcolor}
\definecolor{newcolor}{rgb}{.8,.349,.1}

\usepackage{hyperref}
\usepackage{subcaption} 
\usepackage[switch,pagewise]{lineno}

\usepackage[cachedir=.,frozencache=true]{minted}
\setminted{fontsize=\small,baselinestretch=1}
\usepackage{mathtools}

\newcommand\blfootnote[1]{%
  \begingroup
  \renewcommand\thefootnote{}\footnote{#1}%
  \addtocounter{footnote}{-1}%
  \endgroup
}

\begin{document}

%\verso{Preprint Submitted for review}

%\begin{frontmatter}

% Intersection Count Image?

\title{Radial intersection count image:\\a clutter resistant 3D shape descriptor}

\author{Bart Iver van Blokland \and Theoharis Theoharis}
\date{July 2020}

%\received{1 February 2017}
%\received{\today}

\maketitle

%%%% Do not use the below for submitted manuscripts
%\finalform{28 March 2017}
%\accepted{2 April 2017}
%\availableonline{15 May 2017}
%\communicated{S. Sarkar}

\blfootnote{\copyright 2020. This manuscript version is made available under the CC-BY-NC-ND 4.0 license \url{http://creativecommons.org/licenses/by-nc-nd/4.0/}}

\begin{abstract}
   A novel shape descriptor for cluttered scenes is presented, the Radial Intersection Count Image (RICI), and is shown to significantly outperform the classic Spin Image (SI) and 3D Shape Context (3DSC) in both uncluttered and, more significantly, cluttered scenes. It is also faster to compute and compare. The clutter resistance of the RICI is mainly due to the design of a novel distance function, capable of disregarding clutter to a great extent. As opposed to the SI and 3DSC, which both count point samples, the RICI uses intersection counts with the mesh surface, and is therefore noise-free. For efficient RICI construction, novel algorithms of general interest were developed. These include an efficient circle-triangle intersection algorithm and an algorithm for projecting a point into SI-like ($\alpha$, $\beta$) coordinates. The 'clutterbox experiment' is also introduced as a better way of evaluating descriptors' response to clutter. The SI, 3DSC, and RICI are evaluated in this framework and the advantage of the RICI is clearly demonstrated.
\end{abstract}  

%\begin{keyword}
%% MSC codes here, in the form: \MSC code \sep code
%% or \MSC[2008] code \sep code (2000 is the default)
%\MSC 41A05\sep 41A10\sep 65D05\sep 65D17
%% Keywords
%\KWD Computers and Graphics\sep Formatting\sep Guidelines
%\end{keyword}

%\end{frontmatter}

%\linenumbers

\section{Introduction}

Local shape descriptors have seen extensive use in a wide variety of applications where determining shape correspondences are beneficial or even required. Such applications include registration  \cite{novatnack2008scale} \cite{malassiotis2007snapshots} \cite{yamany2002surface}, shape segmentation \cite{ovsjanikov2011exploration} \cite{hu2012co} \cite{wu2013unsupervised}, and retrieval \cite{3dor.20161082} \cite{3dor.20171051}. 

Many local 3D shape descriptor methods rely on the surfaces present in the volume around a point to compute the degree to which two points are similar. This also makes them susceptible to any unwanted geometry present in the neighbourhood, commonly referred to as \emph{clutter}. For this reason, clutter has been named as a major factor degrading the performance of current descriptors \cite{guo2016comprehensive}. 

The degree to which different descriptors are capable of resisting the negative effects of clutter varies. One classical method which has shown to be significantly resistant to clutter is the Spin Image \cite{johnson1999using} (SI). This descriptor is invariant under rigid transformations, and has been applied successfully for applications such as shape registration \cite{huber2003fully} and facial recognition \cite{kakadiaris2007three}.

In this paper, we present the Radial Intersection Count Image (RICI) combined with a novel distance function. The new descriptor shares the original concept of the Spin Image but is advantageous in terms of its generation speed and clutter resistance. 

In order to show the effectiveness of the RICI, we propose a repeatable experiment aimed at quantifying the effects of clutter on the matching performance of 3D shape descriptors. The main advantage of this evaluation method is that it can be used with datasets of any size, and ensures scenes are cluttered with natural shapes.

In summary, the contributions of this paper are:
\begin{enumerate}
    \item The novel RICI descriptor and an accompanying distance function, capable of resisting clutter.
    \item Algorithms for efficient generation of RICI descriptors, also capable of accelerating SI construction.
    \item The clutterbox experiment for quantifying the effects of clutter.
    %\item The largest quantitative evaluation of the SI, 3DSC, (and RICI) to date
    \item Evidence that the Support Angle filter proposed in the original SI paper does not necessarily improve matching performance.
    \item Freely available GPU implementations for generating and comparing Spin Image, 3DSC, and RICI descriptors, as well as an implementation of the proposed clutterbox experiment.
\end{enumerate}

\section{Background and Related Work}

Numerous local shape descriptors have been proposed to date \cite{guo2016comprehensive}. The Spin Image has been the foundation for a number of methods, which attempt to improve its matching performance or other limitations. Clutter is a major challenge for object descriptors and few methods have addressed it.

\subsection{Spin Images}

The Spin Image \cite{johnson1999using}, originally presented by Johnson et al., is a classic descriptor generated from an oriented point cloud (vertices with position and normal). 

An SI is constructed around an oriented point, the position of which is in this paper referred to as the Spin Vertex $S_v$. The corresponding normal is referred to as the Spin Normal $S_n$. The combined oriented point describes a line, which is called the Central Axis. 

Computing the descriptor involves placing a square plane whose left side is on the Central Axis, with the Spin Vertex at its vertical halfway point. This plane is subsequently subdivided into $(N_{bins} \times N_{bins})$ equivalently sized bins, and rotated for one revolution around the Central Axis. As the plane rotates, the number of point samples intersecting each bin is counted. The descriptor itself is a histogram of the resulting value of each bin, which can be visualised as an image. 

\begin{figure}
    \centering
    \includegraphics[width=4cm]{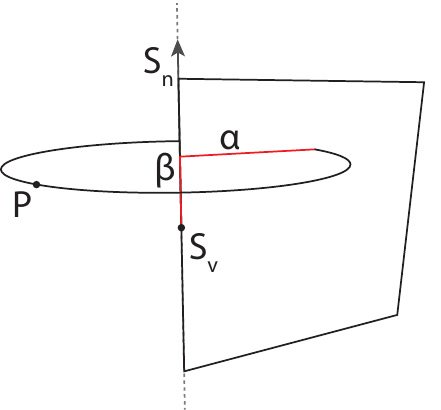}
    \caption{A visualisation of the $\alpha$ and $\beta$ coordinates corresponding to a given point P, relative to the Spin Vertex $S_v$ and Spin Normal $S_n$. The Central Axis; the line described by the Spin Vertex and Spin Normal is also shown.}
    \label{fig:alphabeta}
\end{figure}

In practice, the locations where point samples will intersect with the rotating square can be computed directly as two-dimensional cylindrical coordinates. Here the $\alpha$ coordinate refers to the distance from the point sample to the closest point on the Central Axis, and the $\beta$ coordinate refers to the distance from this closest point to the Spin Vertex. The projection of a given point $P$ is shown in Figure \ref{fig:alphabeta}.

The physical width and height of the square plane is the Support Radius of the descriptor. By rotating the plane around the Central Axis, a cylindrical volume is created, which represents the Support Volume of the descriptor. Additionally, point sample contributions are divided over nearby bins using bilinear interpolation to reduce the effects of aliasing. 

Johnson et al. also describe a prefiltering step called the Support Angle, where a sample oriented point is not included in the computation of the descriptor if the angle between its normal vector and the Spin Normal exceeds a set threshold.

The descriptor's core idea is that a pair of points with identical surfaces surrounding them, and assuming both have been uniformly sampled, will have proportional quantities of projected points in similar locations. Images can thus be compared using statistical correlation.

\subsection{Methods related to the SI}

One of the major issues with the Spin Image is its volatility. Uniform sampling of triangle meshes as well as scans from 3D capture devices are inherently noisy. Carmichael et al. proposed a method to address this by computing the exact area of the support region intersecting each pixel \cite{carmichael1999large}. 

Other methods aim to address specific limitations of the spin image. Assfalg et al. proposed the spin image signature aimed at simplifying the ease of image retrieval from a large database \cite{assfalg2007content}. Dinh et al. aimed at addressing the issue of selecting bin sizes by creating a spin image variant with variable sized histogram bins \cite{dinh2006multi}, although their solution involves the manual setting of parameters.

An alternate spin image variant, proposed by Guo et al. used three spin images per vertex rather than a single one for better matching performance \cite{guo2013trisi}. Accelerating spin image generation using a GPU was first proposed by Davis et al. \cite{davis20083d} \cite{gerlach2011accelerating}. Alternate derivative methods include Spin Contours, proposed by Liang et al. \cite{liang2015geodesic} and colour spin images by Pasqualotto et al. \cite{pasqualotto2013combining}.

\subsection{The 3D Shape Context}

The 3D Shape Context, proposed by Frome et al. \cite{frome2004recognizing}, is a histogram descriptor constructed by accumulating points by their spherical coordinates and distance relative to an oriented reference point in a spherical support region. The support region is divided into $J$ equally spaced spherical wedges, centred around the central axis described by the reference oriented point (similar to the SI). Each wedge is subsequently divided into $K$ elevation divisions. The bin volumes are finally created by the intersection volume of each radial and elevation divisions with the volume bounded by two of $L$ successive spheres with exponentially increasing radii. 

The descriptor has a degree of freedom around the Central Axis, which the Authors solve by generating $J$ different descriptors for each vertex, where each of the wedges has been offset by a multiple of the angle $\frac{2\pi}{J}$. However, due to its self-symmetry, this step is unnecessary for descriptors used for querying. 

%Some testing for clutter resistance is done, although only with two views of the same scene.

\subsection{Other Clutter-Resistant Shape Matching Methods}

Some methods which have been proposed to date, in addition to the Spin Image and 3DSC, have been shown to perform better in cluttered scenes than others \cite{guo2016comprehensive} \cite{mian2006three}. 

Mian et al. presented a method which creates a three-dimensional grid of voxels based on two randomly selected vertices, referred to as a Tensor \cite{mian2006three}. Their results outperform the Spin Image, and show resistance to clutter being present in the scene. % evaluation is somewhat similar to mine, but they too don't sufficiently control the geometric properties variable

The THRIFT descriptor, proposed by Flint et al. \cite{flint2007thrift}, uses an approach similar to the Scale-Invariant Feature Transform (SIFT) by Lowe et al \cite{lowe2004distinctive}. The method aims to find distinctive points which can be detected reliably under a wide range of conditions. This is accomplished by computing a three-dimensional density map of the input point cloud, and selects interest points by locating local maxima of the Hessian matrix. 

Local surface patches, proposed by Chen et al. \cite{chen20073d}, is a two-dimensional histogram descriptor generated from points in an oriented point cloud. Each descriptor accumulates points in a spherical support volume, by their shape index and the cosine of the angles between their normal vectors. The authors only test their method on range images, and do not expose the descriptor to significant levels of clutter themselves. However, experiments performed in the review by Guo et al. \cite{guo2016comprehensive} suggest that this method performs well in cluttered scenes.

Unfortunately, the above works on clutter resistant descriptors used very small datasets for testing their methods (1 to 56 objects). Therefore, the provided results may be statistically biased, since the proposed descriptors were not subjected to a sufficiently wide range of possible surface features. The datasets used were also not made public, making it difficult to compare their results. In addition, some used very similar objects (such as cars), presumably for ease of creation, which is not representative of all forms of clutter that can be encountered in a real scene.

\subsection{Learning Approaches}

More recent shape matching methods have attempted to utilise Neural Networks. One of the major hurdles these methods need to overcome is the inherent irregularity present in 3D shape data, as opposed to more regular data such as images on which learning methods have been applied successfully.

To this end, many methods, such as the PPFNet proposed by Deng et al. \cite{deng2018ppfnet}, make use of existing descriptors or features in a pre-processing step to regularise the input to the neural network. PPFNet specifically uses point pair features, and was shown to outperform many current state-of-the-art handcrafted methods.

Another regularisation approach is the voxelisation of the input point cloud or mesh, which has amongst others been exploited in the 3DMatch method proposed by Zeng et al. \cite{zeng20173dmatch}, who successfully apply their proposed method on point cloud alignment and keypoint matching, outperforming both handcrafted and earlier learning methods. 

While these learning methods show great promise, their applicability depends highly on the used dataset for training, and may require retraining for new environments. Moreover, current learning methods tend to be highly computationally expensive, which can limit their applicability to small datasets only \cite{ioannidou2017deep}.

\section{Radial Intersection Count Images (RICI)}
The novel RICI descriptor is now detailed, which shares some conceptual similarities with the original Spin Image, and has preliminarily been proposed as a quasi Spin Image \cite{vanquasi}.

\subsection{RICI Generation}
\label{sec:projection}

A RICI descriptor is a 2D histogram of integers. It is constructed around an oriented point, and has a Central Axis around which a square plane is conceptually rotated, similar to the Spin Image. The square plane is divided into ($N_{bins} \times N_{bins}$) bins, producing a histogram which can be visualised as a grayscale image.

The primary difference between the RICI and the SI is what is counted in each histogram bin. In Spin Images, projected point samples are accumulated to create an estimate of the surface area intersecting each bin or pixel as the square plane is rotated for a full revolution. In contrast, RICI bins count the number of intersections of circles with the surfaces of the scene and are thus integers.

\begin{figure}
    \centering
    \includegraphics[width=8cm]{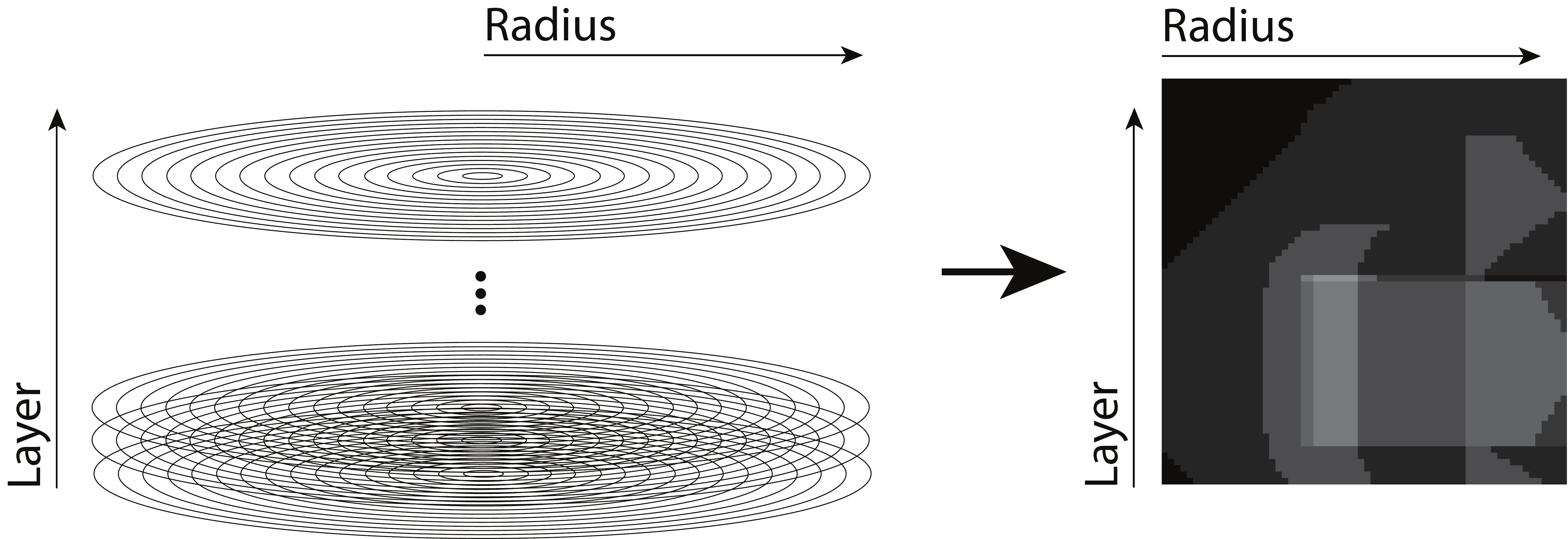}
    \caption{A visualisation of the construction of a RICI image.}
    \label{fig:RICI_stack}
\end{figure}

The conceptual construction method, i.e. the relationship between the aforementioned intersection circles and the produced descriptor is visualised in Figure \ref{fig:RICI_stack}. Consider a set of circles that are centred at fixed distances from the Spin Vertex on the Central Axis and have a fixed number of radii. Each bin in the RICI image stores the number of intersections of the corresponding circle with the surfaces of the scene. RICI rows thus represent circles on the same plane, and RICI columns circles with equivalent radii.

\begin{figure}
    \centering
    \includegraphics[width=5cm]{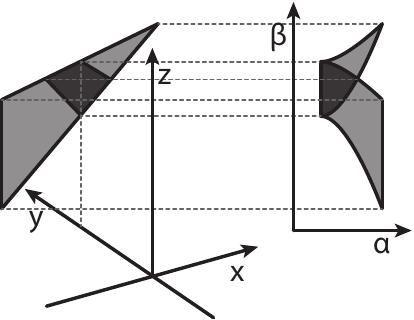}
    \caption{\small A triangle depicted alongside its projection in cylindrical coordinate space. The area in which circles centred and directed along the z-axis intersect the triangle twice is coloured in dark grey. Sizes may not be to scale.}
    \label{fig:triangle_correspondence}
\end{figure}

The remainder of this section presents a method for efficiently computing RICI descriptors. The general idea is to iterate over each triangle in the scene, and determine the set of circles in cylindrical coordinates (see Figure \ref{fig:alphabeta}) which will intersect with it. This implies a complexity of O(T), where $T$ is the number of triangles in the scene, as in the worst case, the number of circles is fixed and equal to the resolution of a RICI image. The bins corresponding to these circles are incremented. Note that cylindrical projections will not preserve the linearity of a triangle's edges (as shown in Figure \ref{fig:triangle_correspondence}), thus not allowing the use of common rasterisation methods. Instead we exploit a circle-triangle intersection algorithm in order to determine the correct projections. 

To summarise, a RICI image is generated by iterating over each triangle in the scene, and in turn each triangle is processed in 3 steps:

\begin{enumerate}
    \item Project the triangle vertices into cylindrical coordinate space, as described in Section \ref{subsubsec:projectionalgorithm}.
    \item Using the circle-triangle intersection method outlined in Section \ref{subsec:intersectionalgorithm}, compute the range of $\alpha$ coordinates which will intersect with the triangle for each $\beta$ coordinate in the triangle's $\beta$-extent. 
    \item Increment the histogram bins that correspond to these intersections.
\end{enumerate}

\subsubsection{Projecting Vertices into Cylindrical Coordinate Space}
\label{subsubsec:projectionalgorithm}

An efficient method for projecting points from Euclidean coordinates into cylindrical coordinates is presented. Apart from the RICI, this method can also be applied directly in the construction of SI descriptors.

The algorithm projects a point $P = (P_{x}, P_{y}, P_{z})$ by computing two transformations. First, a translation that moves the Spin Vertex $S_{v} = (S_{vx}, S_{vy}, S_{vz})$ to the origin (Equation \ref{eq:relativeLocation}), and second, a rotation which aligns the Spin Normal $S_{n} = (S_{nx}, S_{ny}, S_{nz})$ with the z-axis. The projected point's $\alpha$ and $\beta$ coordinates can be computed trivially afterwards.

\begin{figure}
    \centering
    \includegraphics[width=3cm]{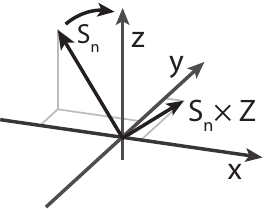}
    \caption{\small Direct approach for vector alignment. First, compute the vector product $S_n \times Z$ between the spin normal $S_n$ and z-axis. Second, rotate $S_n$ around $S_n \times Z$ to align it with the z-axis.}
    \label{fig:directRotation}
\end{figure}

For the z-axis alignment transformation, a common technique for aligning two vectors consists of a vector product followed by a rotation (shown in Figure \ref{fig:directRotation}). While the vector product itself is inexpensive (due to one of the vectors being the z-axis) the subsequent alignment rotation requires a relatively expensive multiplication with a 3x3 matrix. 

\begin{figure}
    \centering
    \begin{subfigure}[t]{4cm}
        \centering
        \includegraphics[width=3cm]{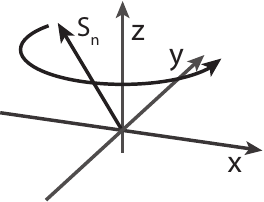}
        \caption{\small Rotation 1: \\Align the spin normal with the XZ-plane by a rotation around the Z-axis (Equation \ref{eq:firstRotation})}
        \label{fig:firstRotation}
    \end{subfigure}
    ~
    \begin{subfigure}[t]{4cm}
        \centering
        \includegraphics[width=3cm]{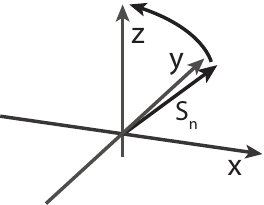}
        \caption{\small Rotation 2: \\Align the spin normal with the Z-axis by a rotation around the Y-axis (Equation \ref{eq:secondRotation})}
        \label{fig:secondRotation}
    \end{subfigure}
    \caption{\small Visual representation of the rotations that form our alignment method.}
    \label{fig:alignmentComparison}
\end{figure}

Our alignment method instead uses two rotations, exploiting the observation that only distance must be preserved for the $\alpha$ coordinate. We align the spin normal with the xz-plane using a rotation around the z-axis (see Figure \ref{fig:firstRotation} and Equation \ref{eq:firstRotation}). We then align the transformed normal with the z-axis by a rotation around the y-axis (Figure \ref{fig:secondRotation} and Equation \ref{eq:secondRotation}).

\begin{equation} \label{eq:rotationCoefficients}
\begin{split}
    [N_{ax}, N_{ay}] &= Normalize[S_{nx}, S_{ny}] \\
    [N_{bx}, N_{bz}] &= Normalize[S_{nx}, S_{nz}]
\end{split}
\end{equation}

\begin{equation} \label{eq:relativeLocation}
\begin{split}
    P'_{x} &= P_{x} - S_{vx} \\
    P'_{y} &= P_{y} - S_{vy} \\
    P'_{z} &= P_{z} - S_{vz} 
\end{split}
\end{equation}

\begin{equation} \label{eq:firstRotation}
    \begin{split}
    P''_{x} &= N_{ax} \cdot P'_{x} + N_{ay} \cdot P'_{y} \\
    P''_{y} &= -N_{ay} \cdot P'_{x} + N_{ax} \cdot  P'_{y}
    \end{split}
\end{equation}

\begin{equation} \label{eq:secondRotation}
    \begin{split}
    T_{x} &= N_{bz} \cdot P''_{x} - N_{bx} \cdot P'_{z} \\
    T_{y} &= P''_{y} \\
    T_{z} &= N_{bx} \cdot P''_{x} + N_{bz} \cdot P'_{z}
    \end{split}
\end{equation}

\begin{equation} \label{eq:calculateAlphaBeta}
    \begin{split}
    \alpha_i &= | (T_{x}, T_{y}) | \\
    \beta_i &= T_{z}
    \end{split}
\end{equation}

The coefficients of the rotation transformations $N_a$ and $N_b$ can be calculated inexpensively from components of the spin normal $S_n$, as shown in Equation \ref{eq:rotationCoefficients}. When both coefficients of either $N_a$ or $N_b$ are zero, that rotation step is unnecessary and an identity rotation is used instead. The key here is that, considering a two-dimensional coordinate system $xy$, the coordinates of a normalised vector represent the sine and cosine values of a rotation which aligns that vector with the x-axis. These normalised coordinates can therefore be used directly for this purpose. 

It should be noted that since the rotation coefficients only depend on the spin normal, they are constant for the entire spin image. Therefore they only need to be computed once per image, essentially taking this computation out of the inner loop. This is the primary reason for the method's efficiency compared to previous work.

\subsubsection{Circle-Triangle Intersection}
\label{subsec:intersectionalgorithm}

A circle-triangle intersection test can result in four outcomes; no intersection, one intersection, two intersections, or infinite intersections. However, due to floating point rounding errors, handling the latter, while possible, is not feasible in practice and is thus not addressed by the proposed algorithm.

Our algorithm starts off with the triangle vertices in cylindrical coordinate space. For a given $\beta$ coordinate, it determines the range of $\alpha$ coordinates which result in a single or double intersection. This information is subsequently used to ``rasterise'' a row of pixels for the triangle in the RICI descriptor.

\begin{figure}[!t]
    \centering
    \includegraphics[width=3.3in]{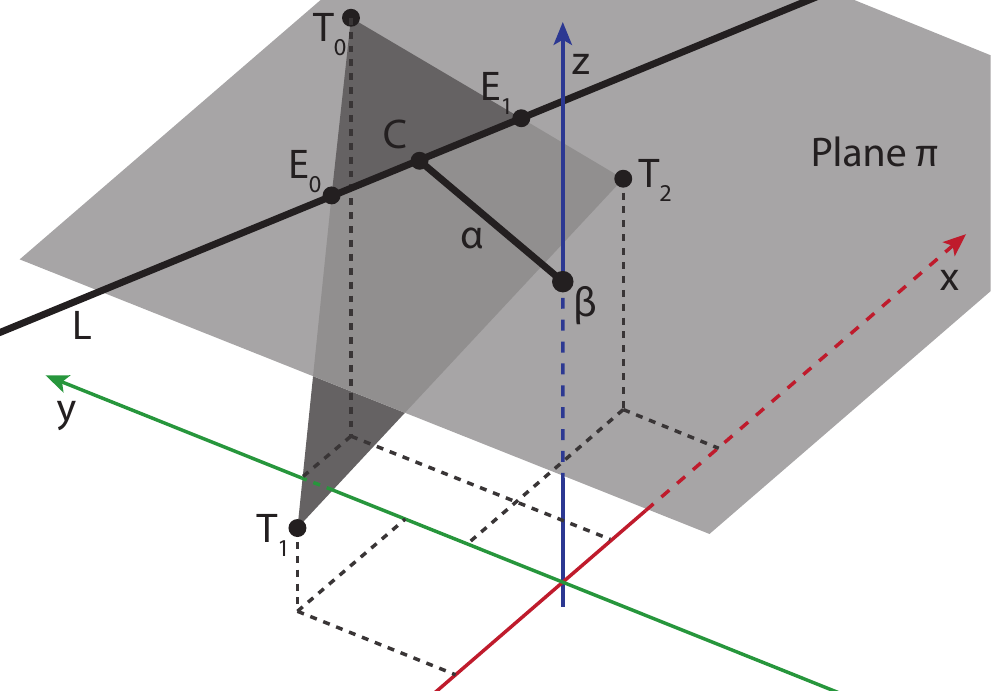}
    \caption{\small A triangle defined by the vertices $T_0$, $T_1$, and $T_2$ intersecting with the horizontal plane through an arbitrary coordinate $\beta$ on the z axis.}
    \label{fig:alignment_intersection}
\end{figure}

The method operates in three distinct stages. First, the triangle is intersected with the plane $\pi$ of the circle, which is parallel to the $xy$ plane, as shown in Figure \ref{fig:alignment_intersection}. Next, the triangle vertices are rotated around the z-axis in order to further simplify subsequent computations. Finally, the ranges of circle radii in which respectively single and double intersections occur, are calculated.

Prior to detailing these stages individually, we will outline the geometric background used in the intersection test calculations.

Figure \ref{fig:alignment_intersection} shows a given $\beta$ coordinate. The triangle being tested is defined by its transformed vertices $T_0$, $T_1$, and $T_2$, using the previously described alignment transformation. Here all points with equal $\beta$ coordinates lie on the plane $\pi$. 

Where the triangle intersects the plane, it forms an intersection line segment $E_0E_1$, which defines a line $L$. The range of $\alpha$ coordinates either intersecting the triangle once or twice can be calculated by determining which radii intersect with this line segment. This reduces the determination of intersection distances to a two-dimensional problem. 

For single intersections, the lower and upper bounds of radii is $[min(|E_0|, |E_1|), max(|E_0|, |E_1|)]$. Note that the 2D coordinates of $E_0$ and $E_1$ are equivalent to the vectors $\vec{\beta E_0}$ and $\vec{\beta E_1}$, respectively.

A double intersection occurs when the closest point to $\beta$ on line $L$ is also on the line segment $E_0$ $E_1$. When double intersections exist, the range of radii in which they occur is $[|C|, min(|E_0|, |E_1|)]$.

\begin{figure}[!t]
    \centering
    \includegraphics[width=3.3in]{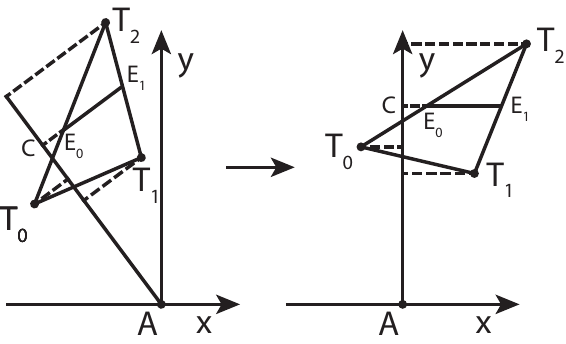}
    \caption{\small Aligning an $\vec{E_{0}E_{1}}$ vector with the x-axis. Any value of $C$ can be chosen for which an $\vec{E_{0}E_{1}}$ vector exists for this purpose. A sample $\vec{E_{0}E_{1}}$ vector has been indicated in the Figure. Point $A$ represents the point on the Central Axis marked by $\beta$ in Figure \ref{fig:alignment_intersection}.}
    \label{fig:alignment_2d}
\end{figure}

Given the aforementioned background, the next step of our method is aligning the vector $\vec{\beta C}$ with the y-axis, as illustrated in Figure \ref{fig:alignment_2d}. The objective of this step is to simplify the remaining calculations for the intersection test. Alignment is done by normalising the vector between $E_0$ and $E_1$, and subsequently rotating the triangle vertices around the z-axis; the coordinates of the normalised vector can be used directly as sine and cosine coefficients for the rotation. 

At this stage, determining the existence of a double intersection is inexpensive, and can be achieved by comparing signs of the $x$ components of the aligned $E_0$ and $E_1$ coordinates. Different signs indicate that a double intersection exists. If so, the length of $\vec{\beta C}$ (the rotated y-coordinate of $C$) represents the lower bound of radii which correspond to double intersections.

The intersection test itself can be done by comparing a given radius against the computed ranges, which yields an intersection count corresponding to that radius.

Summarising, computing the range of values of $\alpha$ that will result in a single or double intersection for a given value of $\beta$ involves the following steps:

\begin{enumerate}
    \item Determine the intersection points $E_0$ and $E_1$ for any value of value of $\beta$ where $L$ is defined, as shown in Figure \ref{fig:alignment_intersection}.
    \item Rotate $E_0$ and $E_1$ around the z-axis such that the vector $\vec{E_{0}E_{1}}$ is aligned with the x-axis (as shown in Figure \ref{fig:alignment_2d}).
    \item Determine the distance of $E_0$ and $E_1$ from the z-axis.
    \item The range of circle radii in which single intersections occur is $[min(|E_0|, |E_1|), max(|E_0|, |E_1|)]$.
    \item Determine the existence of a double intersection by comparing the signs of the x-coordinates of $E_0$ and $E_1$. If they are different then a double intersection exists.
    \item If a double intersection exists, the range of $\alpha$ coordinates (circle radii) corresponding to the double intersection is the y-coordinate of either $E_0$ or $E_1$ and the shortest distance between the z-axis and $E_0$ or $E_1$.
\end{enumerate}

\subsection{A Clutter-Resistant RICI Distance Function}

Spin Images, by their nature of being generated from oriented point clouds, are inherently noisy. They have as such relied on statistical correlation to compute similarity. The idea here is that two matching bins tend to have proportionally similar accumulated sample counts. Unfortunately, this method is susceptible to the effects of clutter. Additional geometry present in the support volume causes portions of the image to receive additional projected point samples, which consequently negatively affects the computed correlation value.

When it comes to comparing RICIs, one important downside of the Pearson Correlation Coefficient is that it is not defined for sequences of constant values. While this scenario is unlikely to occur for Spin Images, there exist situations in which RICIs consist solely of pixels with equivalent intersection counts. For these situations, the Pearson correlation coefficient is undefined, and therefore an insufficient solution for comparing RICIs. Handling these edge cases separately is possible, but results in a solution that requires balancing awarded scores against normal situations.

Meanwhile, the RICI does not have the aforementioned issue of noise, and is as such not bound solely to using statistical methods for measuring similarity. For these reasons we propose a new distance function, which is by design able to resist some of the negative effects of clutter, primarily by exploiting features of the RICI.

\begin{figure}
    \centering
    \begin{subfigure}[t]{4cm}
        \centering
        \includegraphics[width=4cm]{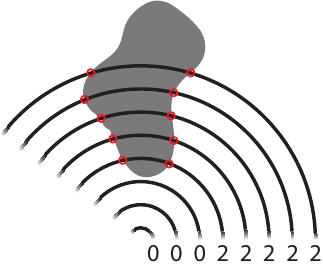}
        \caption{\small Intersection counts without clutter}
        \label{fig:intersectioncountsnoclutter}
    \end{subfigure}
    ~
    \begin{subfigure}[t]{4cm}
        \centering
        \includegraphics[width=4cm]{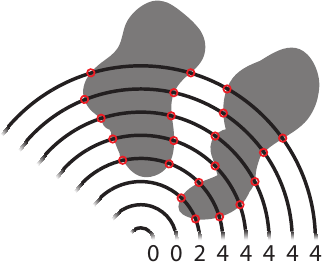}
        \caption{\small Intersection counts with clutter}
        \label{fig:intersectioncountscluttered}
    \end{subfigure}
    \caption{\small Demonstration of changes in intersection counts generally being unaffected by clutter. A portion of a single layer of intersection circles is shown. Intersections with the shape surface have been marked.}
    \label{fig:intersectioncounts}
\end{figure}

First, the distance function does not consider the values of pixels in the RICI. Instead, \emph{changes} in pixel values (i.e. intersection counts which show up as edges in the RICI) are compared. As RICIs are free of noise, it is possible to interpret pixel values directly. The main advantage of this approach is that changes in intersection counts are largely unaffected by clutter. The reason for this can be seen in Figure \ref{fig:intersectioncounts}. 

In Figure \ref{fig:intersectioncountsnoclutter}, a cross section is shown of an arbitrary 3D shape. On the same plane, circles are drawn with increasing radii, similar to how RICI images are computed. The numbers below each circle indicate the number of intersections they encounter, which corresponds to the value of their respective pixels in the RICI image. 

Similarly, Figure \ref{fig:intersectioncountscluttered} shows the same situation in which a clutter object has been added. From the intersection counts can be seen that even though the absolute intersection counts have now changed, the change in intersection counts from the third to the fourth circle, caused by the original object, is still present. 

Second, when searching, our distance function treats the \textit{needle} (query) and the \textit{haystack} image asymmetrically, in contrast to the Pearson correlation coefficient. One can use the needle image to deduce what features to look for in a given haystack image. 

This asymmetry consists of only computing a sum of squared differences distance on pixels where there are \textit{changes} in the needle RICI image.

\begin{listing}
\begin{minted}{python}
def clutterResistantDistance(needle, haystack):
  score = 0
  for row r in [0..N_bins]:
    # Skip first column
    for column c in [1..N_bins]:
      needleDelta = 
            needle[r][c] - needle[r][c-1]
      haystackDelta = 
            haystack[r][c] - haystack[r][c-1]
      if needleDelta != 0:
        score += 
            (needleDelta - haystackDelta) *
            (needleDelta - haystackDelta)
  return score
            
\end{minted}
\caption{\small Pseudocode for our proposed method for computing the distance between two RICI images.}
\label{listing:comparison_algorithm}
\end{listing}

Returning to Figure \ref{fig:intersectioncounts}, we'll assume that Figure \ref{fig:intersectioncountsnoclutter} shows a cross section of the needle object that we are attempting to locate in the cluttered haystack scene shown in Figure \ref{fig:intersectioncountscluttered}. In our needle image, only the increased intersection counts from the third to the fourth circle are relevant. Including other pixels is not relevant, as there are no changes in the needle image's intersection counts. We can therefore ignore these pixels in our distance computation. This also means any clutter present in the haystack image is ignored by this method.

The proposed Clutter Resistant Distance function $CRD(needleRICI, haystackRICI)$ is shown in Equation \ref{eq:distanceFunctionExpression}, and the corresponding pseudocode is given in Listing \ref{listing:comparison_algorithm}. Note here that the distance function is positive, but not symmetric. It has a complexity of O(1), because comparing a descriptor pair requires a fixed number of operations.

\begin{align}
\begin{split}
D(rici, r, c) &= rici(r,c) - rici(r,c-1)
\end{split}\\
\scriptsize
\begin{split} \label{eq:distanceFunctionExpression}
CRD(n, h) &= \sum_{r = 0}^{N_{bins}}  \sum_{c = 1}^{N_{bins}} \begin{cases} (D(n,r,c) - D(h,r,c))^{2},& \text{if } D(n,r,c)\neq 0\\ 0,& \text{otherwise} \end{cases}
\end{split}
\end{align}
\normalsize

\section{Evaluation}
\label{sec:evaluation}
%It is worth noting that the built-in rasterisation hardware of the GPU could not be utilised for this purpose due to the non-linearity of cylindrical coordinates. An example of this can be seen in Figure \ref{fig:triangle_correspondence}.

The proposed method has been evaluated in terms of its clutter resistance, generation speed, and matching performance. Where applicable, we compare our method against the two most referenced among those listed in survey \cite{guo2016comprehensive} as being clutter resistant. These are the Spin Image\footnote{\cite{tombari2010unique} and \cite{guo2013rotational} also support the SI as a clutter resistant descriptor.} and the 3D Shape Context. It is worth noting that the survey also observes that popular descriptors such as the Fast Point Feature Histogram \cite{Rusu_ICRA2011_PCL}, Unique Signatures of Histograms \cite{tombari2010unique}, and Rotational Projection Statistics \cite{guo2013rotational}, do not exhibit optimal performance under cluttered conditions. We have therefore implemented the above two most referenced clutter resistant methods on the GPU, to allow a direct comparison on the same dataset. 

The novel Clutterbox Experiment is proposed in order to evaluate the effect of clutter on the descriptors' matching performance.

\subsection{The Clutterbox Experiment}

In previous work, clutter has typically been defined as the proportion of area within the support volume that does not belong to the object being recognised. Greater proportions of clutter generally imply worse descriptor performance. The expression used in previous work, initially proposed by Johnson et al. \cite{johnson1999using} is shown in Equation \ref{eq:clutter}. Here $A_{all}$ is the surface area of all objects within the support volume and $A_{object}$ is the surface area of the object of interest.

\begin{equation}
\label{eq:clutter}
    clutter = \frac{A_{all} - A_{object}}{A_{all}}
\end{equation}

The objective of the proposed evaluation method, which we call the ``clutterbox experiment'', is to measure the relationship between increasing levels of clutter and the resulting performance of the descriptor being tested.

In previous clutter experiments, clutter has generally been evaluated by measuring descriptor performance against levels of clutter present at points in a scene without controlling the points' identities. However, this measures the effects of two parameters combined; the descriptor's ability to recognise the desired shape, and the level of clutter present around it. Ideally an evaluation of the effects of clutter should control the former of these parameters, while varying the latter. This is the primary objective that the clutterbox experiment addresses.

Varying clutter levels in the neighbourhood of an object can be done trivially by adding triangles, points, spheres, or cubes in random locations and sizes around an object. However, this kind of clutter is not representative of the clutter that can be expected in a realistic 3D scene. The clutterbox experiment therefore inserts complete objects rather than random noise. This results in a more natural distribution of clutter in the scene, and therefore more directly measures the effect of clutter that can be expected of a given descriptor when applied in a practical context.

The clutterbox experiment is executed a large number of times by varying objects and their transformations, in order to provide robust results, independent of object type.

The steps of the experiment are outlined below:

\begin{enumerate}
    \item Define the clutterbox as a cube of side $s$.
    \item Select $n$ objects at random from a large object collection.
    \item Scale and translate each object such that it fits exactly inside a unit sphere.
    \item Pick one of the $n$ objects at random. This is the reference object.
    \item Compute the reference descriptor set $\{RD\}$, by computing one descriptor for each unique vertex of the reference object.
    \item For each of the $n$ objects in random order, but starting with the reference object: 
    \begin{enumerate}
        \item Place the object within the clutterbox, at a randomly chosen orientation and position, with the constraint that the bounding sphere fits entirely within the clutterbox.
        \item Compute the set of cluttered descriptors $\{CD\}$, by computing one descriptor for each unique vertex of the combined mesh in the clutterbox.
        \item For each $d\in\{RD\}$, create a list of ranked distances to all $c\in\{CD\}$. Keep the rank where the corresponding cluttered descriptor was found in the ranked list ($0 \leq rank \leq |\{CD\}|-1$). Note that lower ranks are better.
        \item Create a histogram where bin $i$ holds the number of times the correct vertex is found in the search results at rank $i$.
    \end{enumerate}
\end{enumerate}

\begin{figure*}
    \centering
    \begin{subfigure}[t]{3.2cm}
        \includegraphics[width=3.2cm]{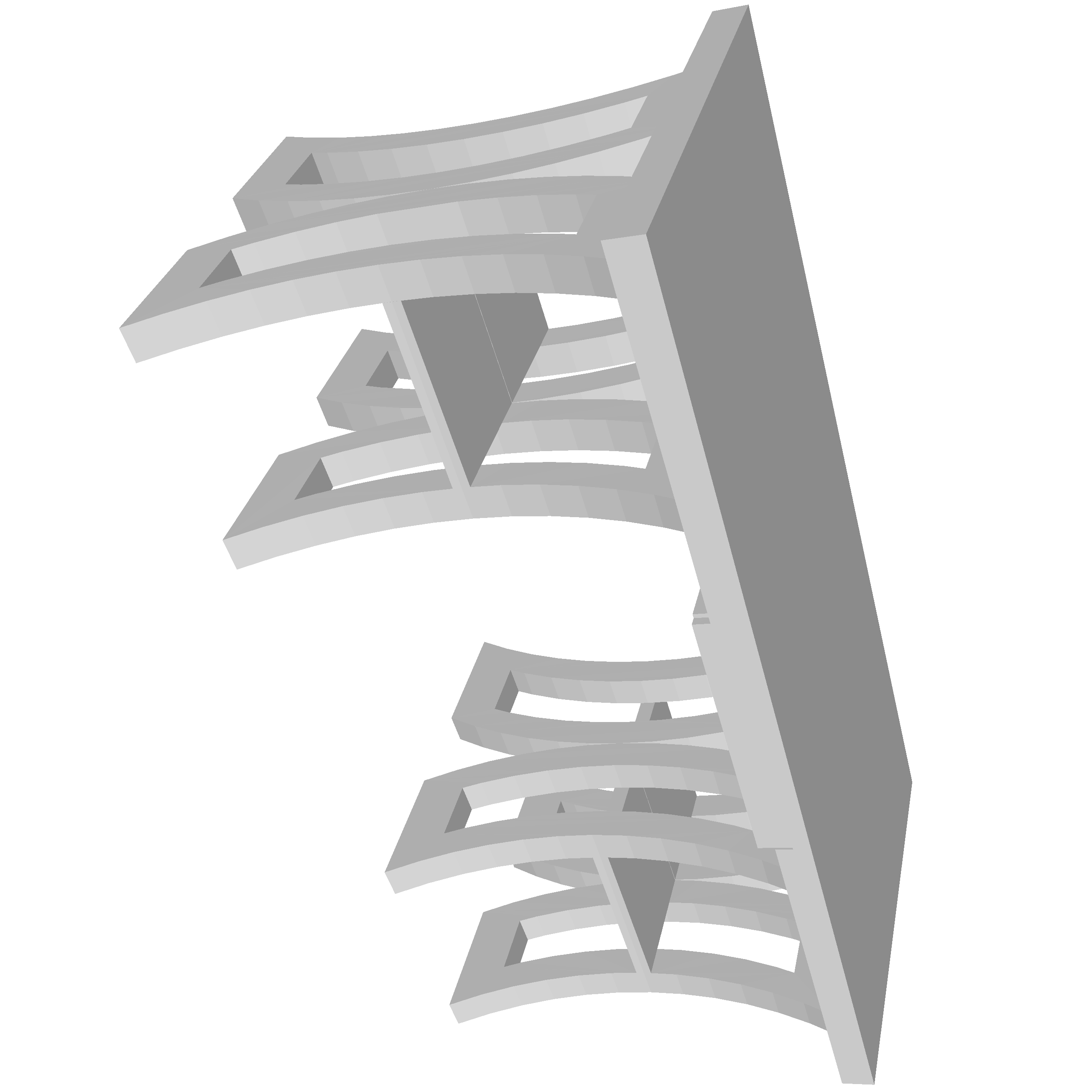}
    \end{subfigure}
    \begin{subfigure}[t]{3.2cm}
        \includegraphics[width=3.2cm]{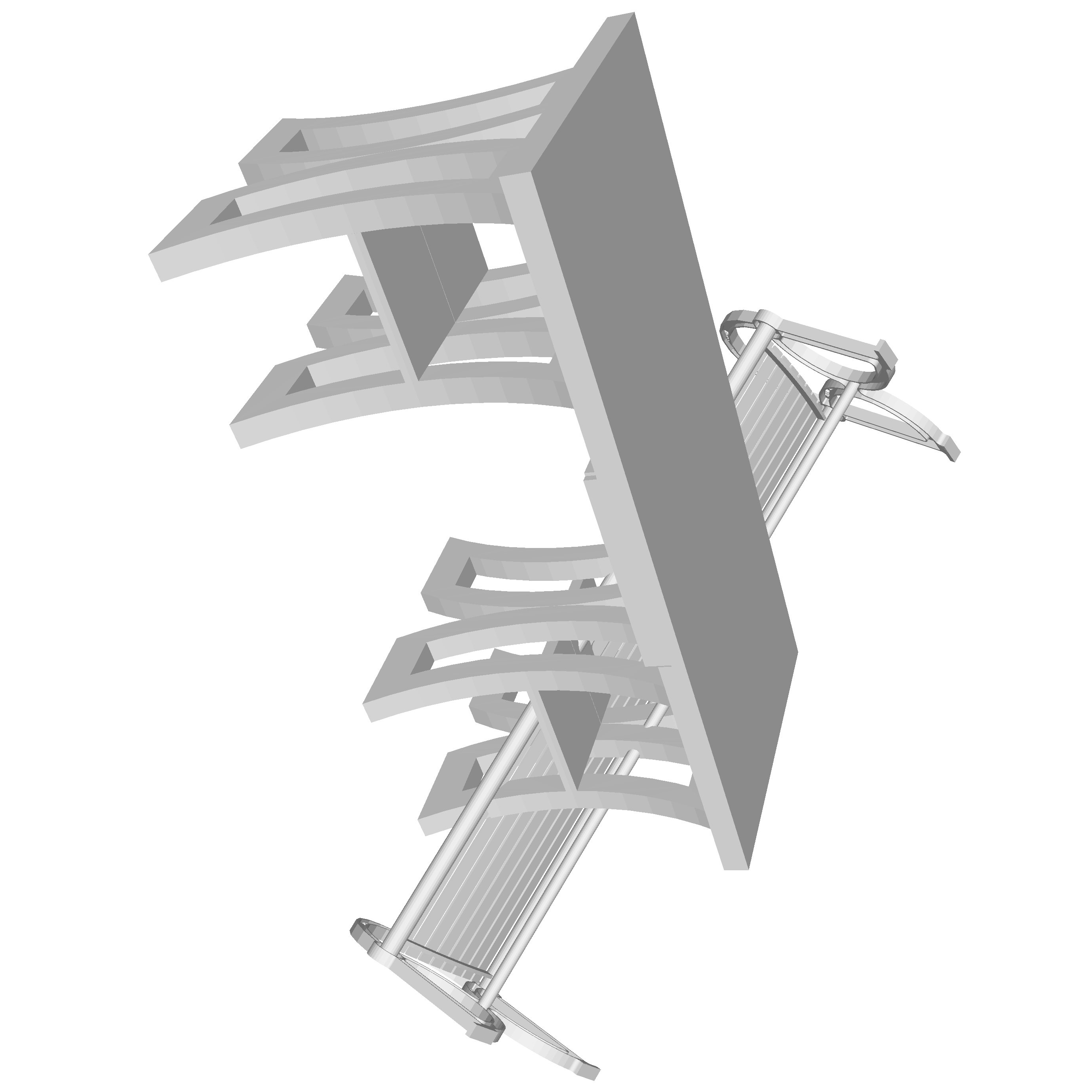}
    \end{subfigure}
    \begin{subfigure}[t]{3.2cm}
        \includegraphics[width=3.2cm]{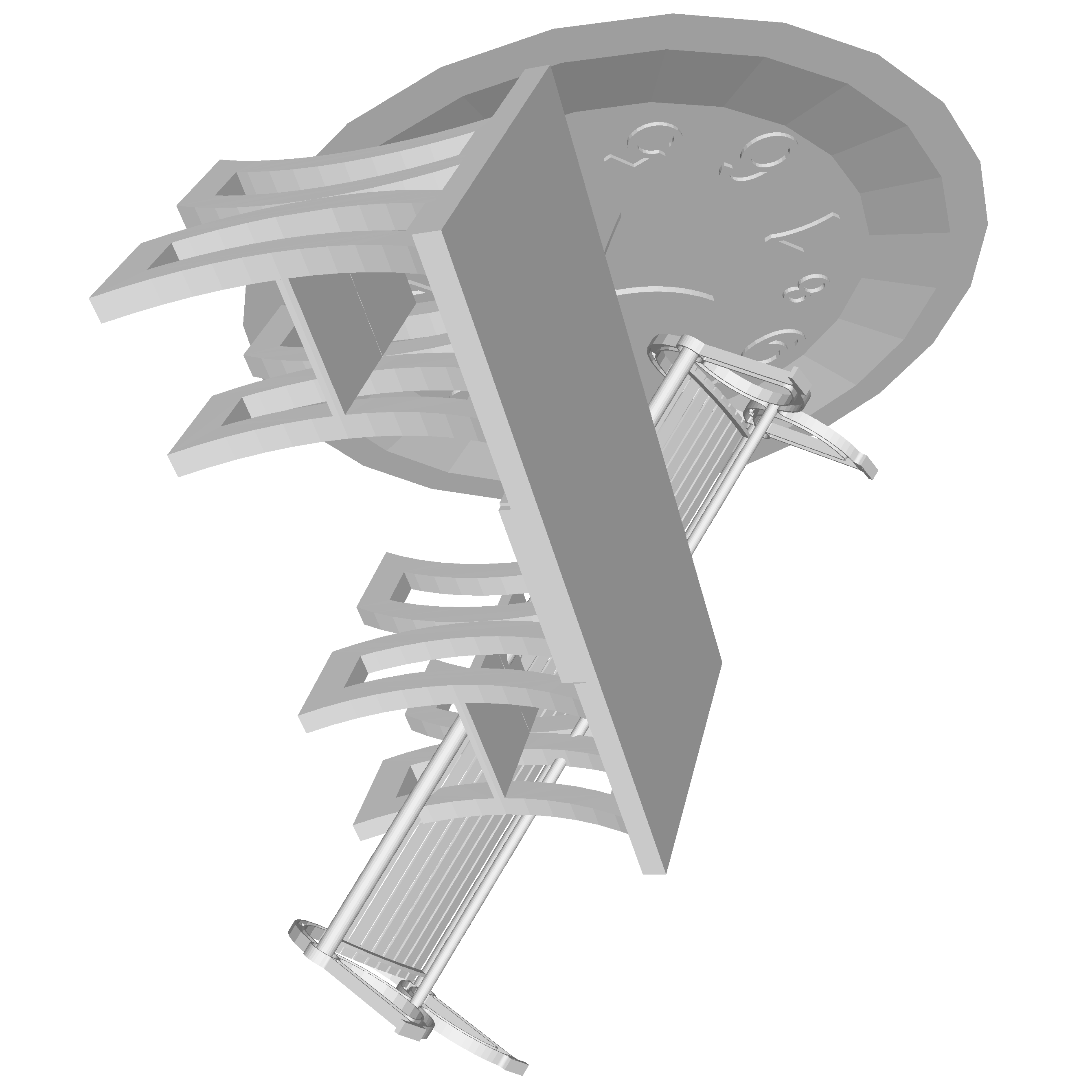}
    \end{subfigure}
    \begin{subfigure}[t]{3.2cm}
        \includegraphics[width=3.2cm]{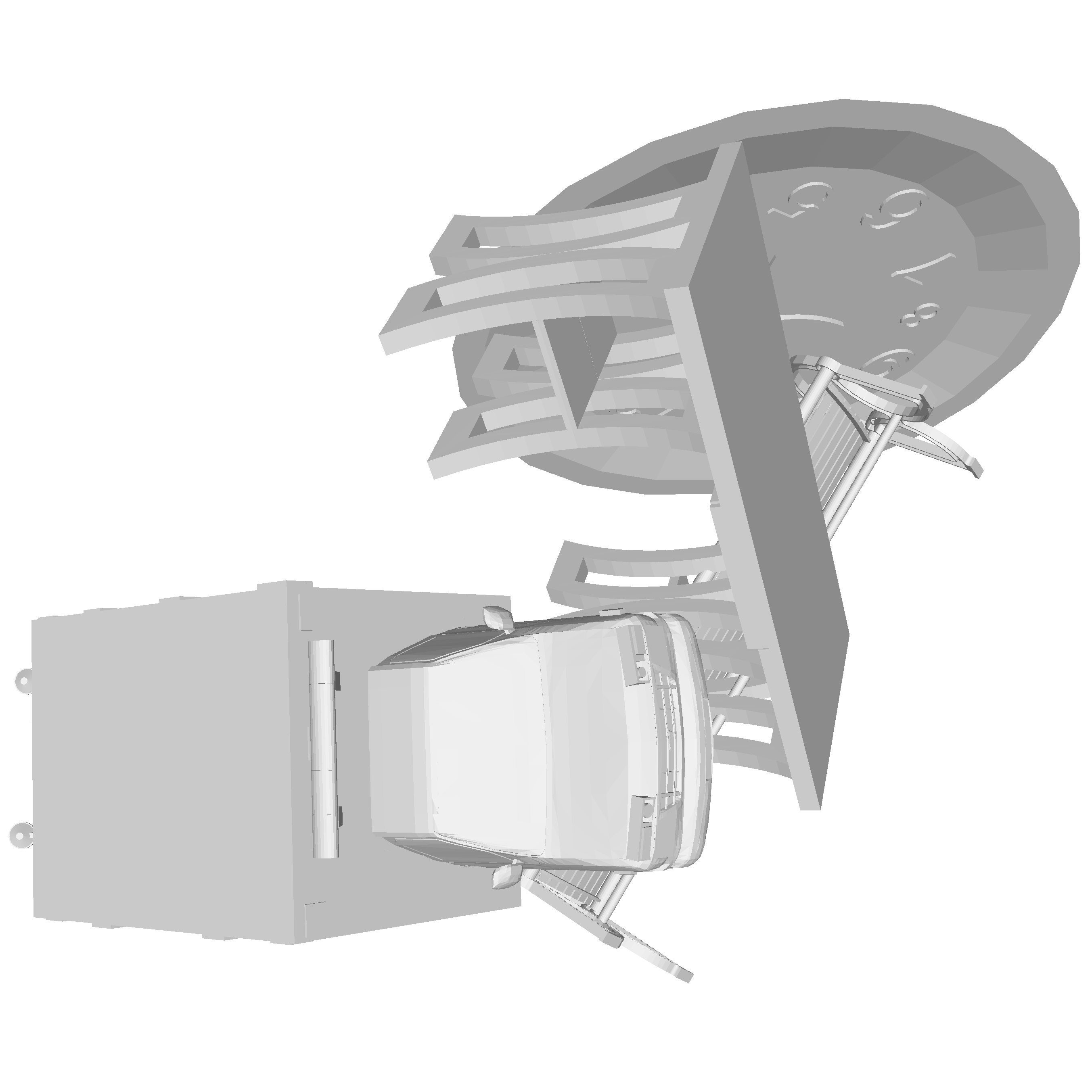}
    \end{subfigure}
    \begin{subfigure}[t]{3.2cm}
        \includegraphics[width=3.2cm]{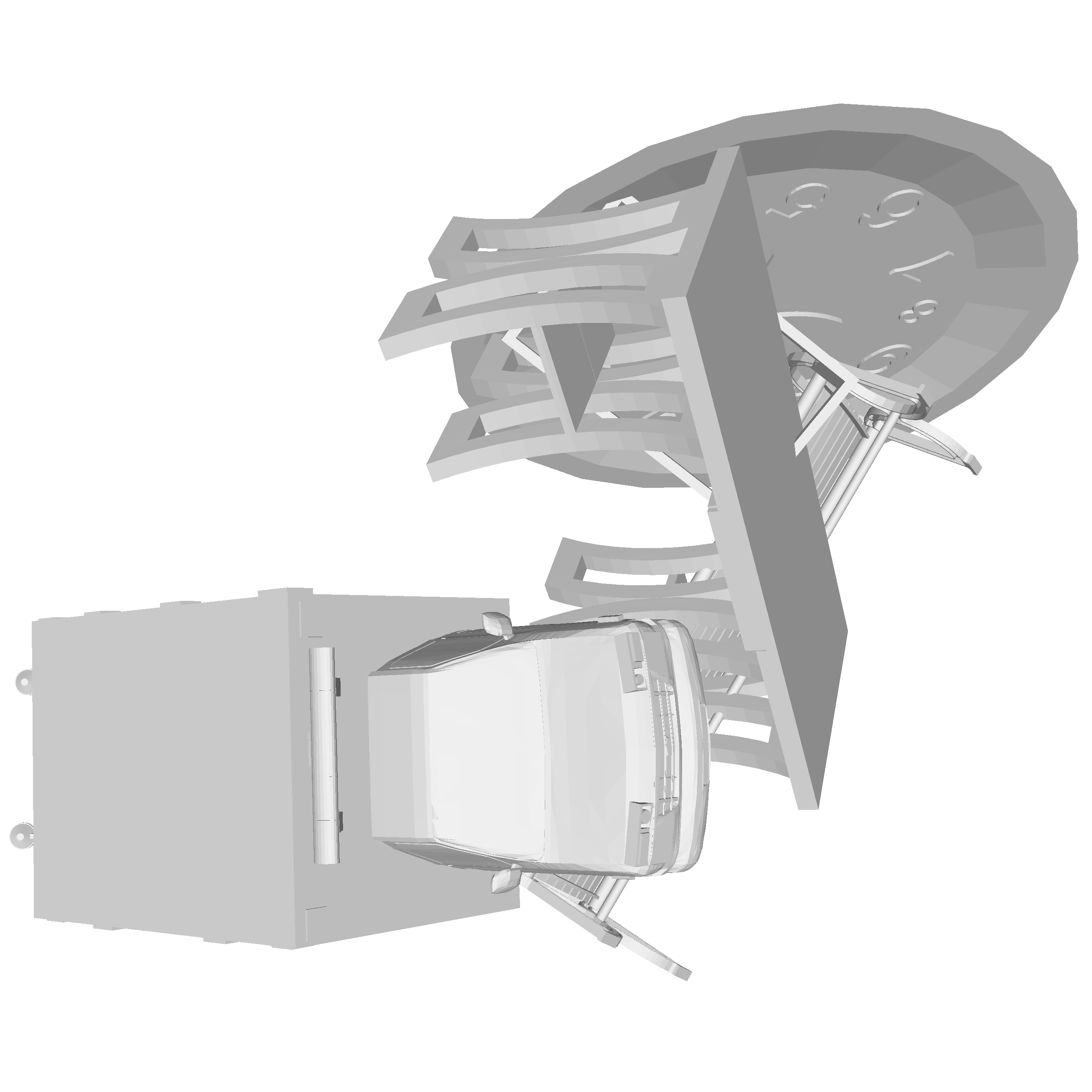}
    \end{subfigure}

    \caption{\small Visual representation of the increasing number of clutter objects added into the clutterbox. The leftmost image only contains the reference object.}
    \label{fig:clutterbox}
\end{figure*}

Thus the output of the clutterbox experiment is a list of histograms, one for each level of clutter. A visualisation of a sequence of scenes with increasing clutter generated by the above experiment is shown in Figure \ref{fig:clutterbox}. 

\subsection{Clutter Resistance Evaluation}

We used the clutterbox experiment to quantify the effects of clutter on the SI and 3DSC versus the proposed RICI descriptor. For our object collection, we selected the combined SHREC2017 dataset \cite{savva2017shrec}, which consists of 51,162 triangle meshes.

In the case of the SI and 3DSC, the combined triangle mesh of the reference and clutter objects was sampled into a point cloud before generating their descriptors; RICI descriptors are generated from the triangle mesh directly. For optimal performance, SI and 3DSC require a high number of samples to ensure a low level of noise in the produced descriptors. However, one cannot increase the sample count indefinitely as that results in a lower generation rate. Based on our experimental evidence on the given dataset, we feel that 10 samples per triangle is a reasonable point on this trade-off.

While Johnson et al. define the bin size (thus the support radius) of the SI to be equal to the mesh resolution, we do not believe their reasoning holds any longer for present day 3D objects. Similar objects can have significant variance in their resolution. As such, making the support radius dependent on the mesh resolution is not a guarantee for better matching performance. We therefore use a constant support radius for all tested methods, set to 0.3 units, relative to the bounding unit sphere, for all scenes in the experiment for ease of reproducibility. For the 3DSC, we set the minimum support radius to $r_{min} = 0.048$ units, which is proportionally the same as the one originally used by Frome et al. \cite{frome2004recognizing}. 

\begin{figure}%[H]
    \centering
    \includegraphics[width=8cm]{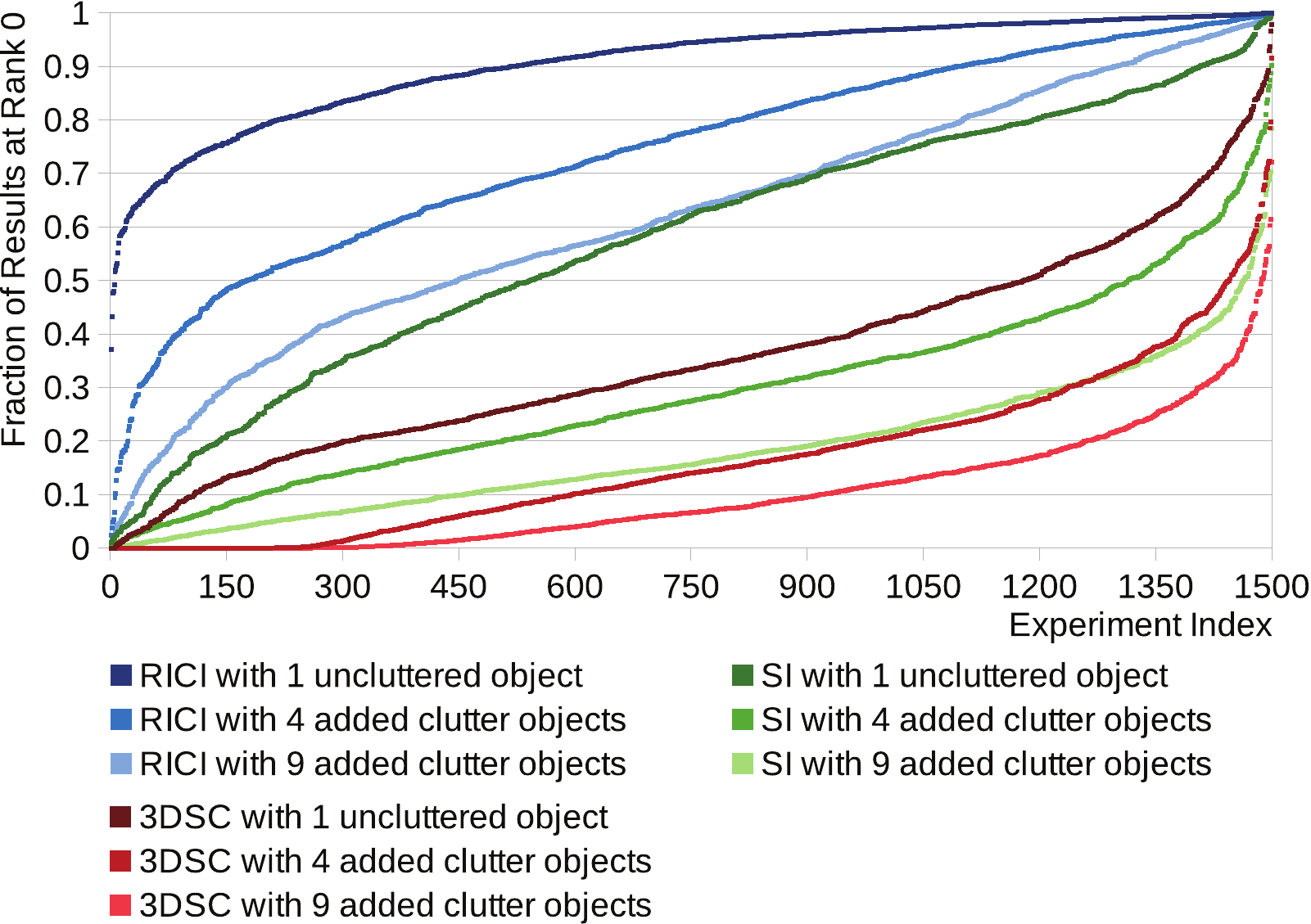}
    \caption{\small Percentage of search results for all tested methods that ended up at rank 0 for each of the 1500 performed experiments.}
    \label{fig:results_RICI_vs_si}
\end{figure}

We executed the experiment 1,500 times, iteratively cluttering a scene with $n = 1$ (the reference object only), $n = 5$, and $n = 10$ objects, into a clutterbox of size $s = 3$. The size of the RICI and SI descriptors $N_{bins}$ was set to 64x64 bins, while the 3DSC descriptor's dimensions were left the same as those used in previous work ($J = 15$, $K = 11$, $L = 12$ \cite{guo2016comprehensive} \cite{frome2004recognizing}). A more detailed discussion on size settings can be found in Section \ref{subsec:settings_discussion}. In order to visualise the histograms generated by the clutterbox experiment, we opted to compute the fraction of the bin representing rank 0 in the histogram against the sum of all bins (all search results). For clarity,  each sequence of such fractions has been sorted individually to produce monotonically increasing curves. The results are shown in Figure \ref{fig:results_RICI_vs_si}.

\begin{figure}%[H]
    \centering
    \includegraphics[width=8cm]{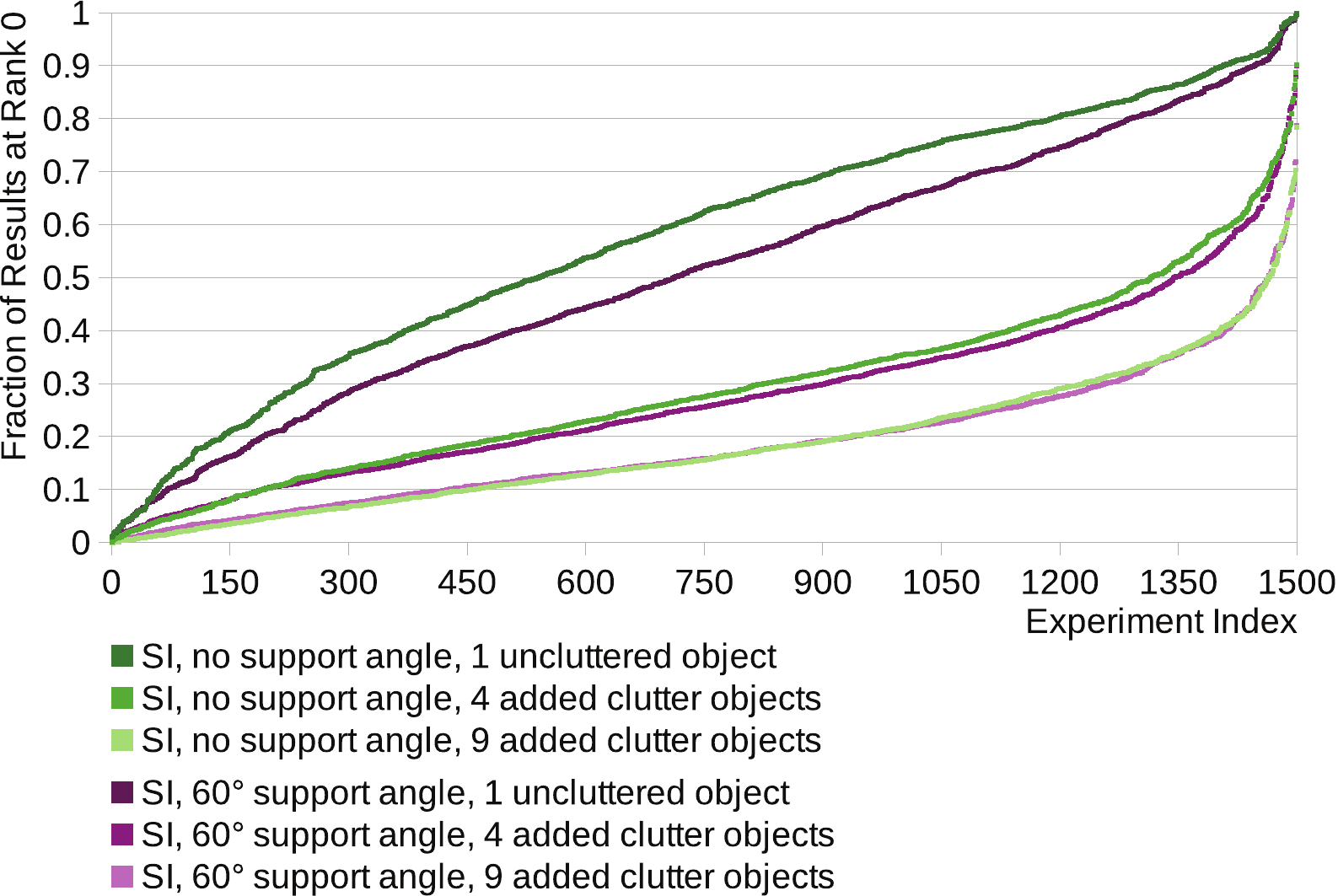}
    \caption{\small Percentage of SI search results that ended up at rank 0 for each of the 1500 performed experiments for two different support angles.}
    \label{fig:results_support_angle}
\end{figure}

The support angle parameter used to generate the SI results in Figure \ref{fig:results_RICI_vs_si} requires further elaboration. In their original SI paper, Johnson et al. claim this filter reduces the effects of self-occlusion and clutter. However, our testing which compared using a support angle filter to not filtering any input points (Figure \ref{fig:results_support_angle}) could not confirm this. All SI results in this paper therefore do not apply any support angle filter, as this favours the SI.

\begin{figure*}
    \centering
    \begin{subfigure}[t]{16cm}
        \includegraphics[width=16cm]{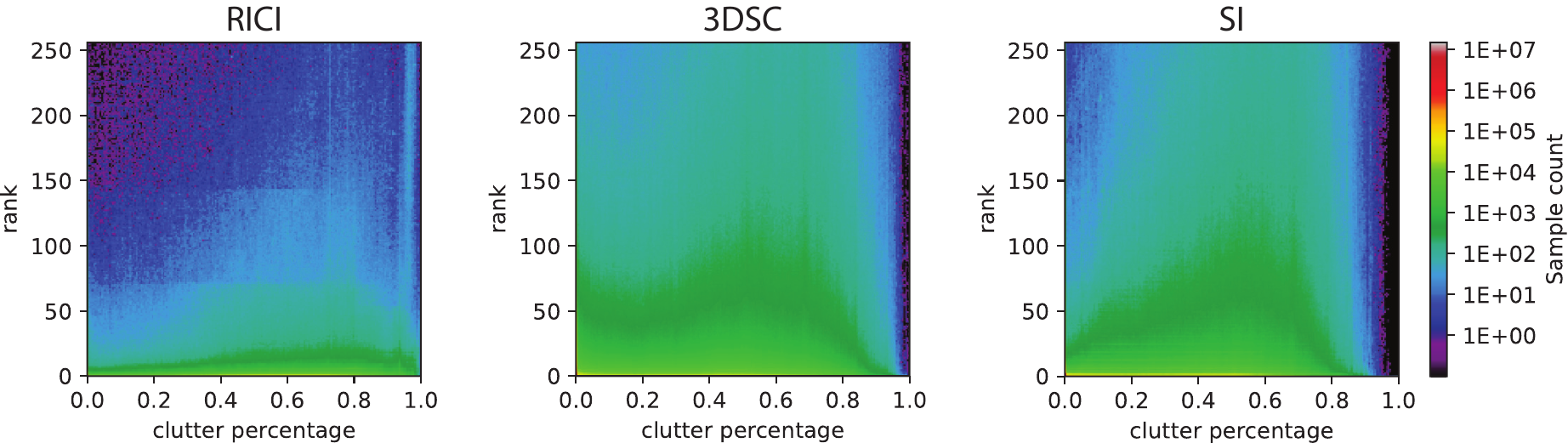}
    \end{subfigure}
    \caption{\small Visualisation of the clutter resistance of RICI, SI, and 3DSC. Colours are mapped using a logarithmic function (colours toward the red end of the spectrum lower in the images is better). A pixel's colour represents the number of search results, i.e. descriptors, that ended up in the specific rank in relation to the amount of clutter within their support volume.}
    \label{fig:clutterComparison}
\end{figure*}

While Figure \ref{fig:results_RICI_vs_si} shows that our RICI descriptor clearly outperforms both the SI and 3DSC in scenes that contain clutter (see Equation \ref{eq:clutter}), it is also relevant to gain insight in the relationship between descriptor performance and the specific clutter level present in the support region. Figure \ref{fig:clutterComparison} shows a heatmap plot of the fractional area of clutter present in the support volume around each Spin Vertex, versus the rank of the corresponding descriptor in the haystack. It can be observed that the RICI trends towards lower ranks than the SI and 3DSC, even at high levels of clutter. Furthermore, while the 3DSC generally does not outperform the SI, it appears more clutter resistant than the SI at extreme clutter levels ($>$ 90\%).

The heatmaps have been computed over 73.5 million search results extracted from scenes with 4 added clutter objects, based on the results of the Clutterbox experiment.

It is not expected that a RICI image would be very dependent on mesh resolution (which may be related to scanning) as intersection counts should in most cases not be very sensitive to that. 

%It is expected that the quality of the presented results is affected in cases where the mesh resolution varies to such a degree that changes in intersection counts are on average displaced by the distance between two neighbouring circles. However, the extent of this has not been investigated as part of this evaluation.

The experiment was implemented using C++, with the descriptor generation and search kernels written in CUDA 10.0. The code was written in such a way that given a dataset of objects, a single random seed determines all randomly chosen parameters, making all results reproducible. The experiment was executed on a combination of Nvidia Tesla cards (P100 16GB, V100 16GB, and V100 SXM3 32GB). All time-based results were exclusively gathered on the latter. One relevant implementation detail is that in cases where multiple search results have the same distance (which may occur due to reasons such as object self-similarity), we use the highest (best) rank of the matched haystack image for the sake of consistency. 

\subsection{Generation Performance}

\begin{figure}%[H]
    \centering
    \includegraphics[width=8cm]{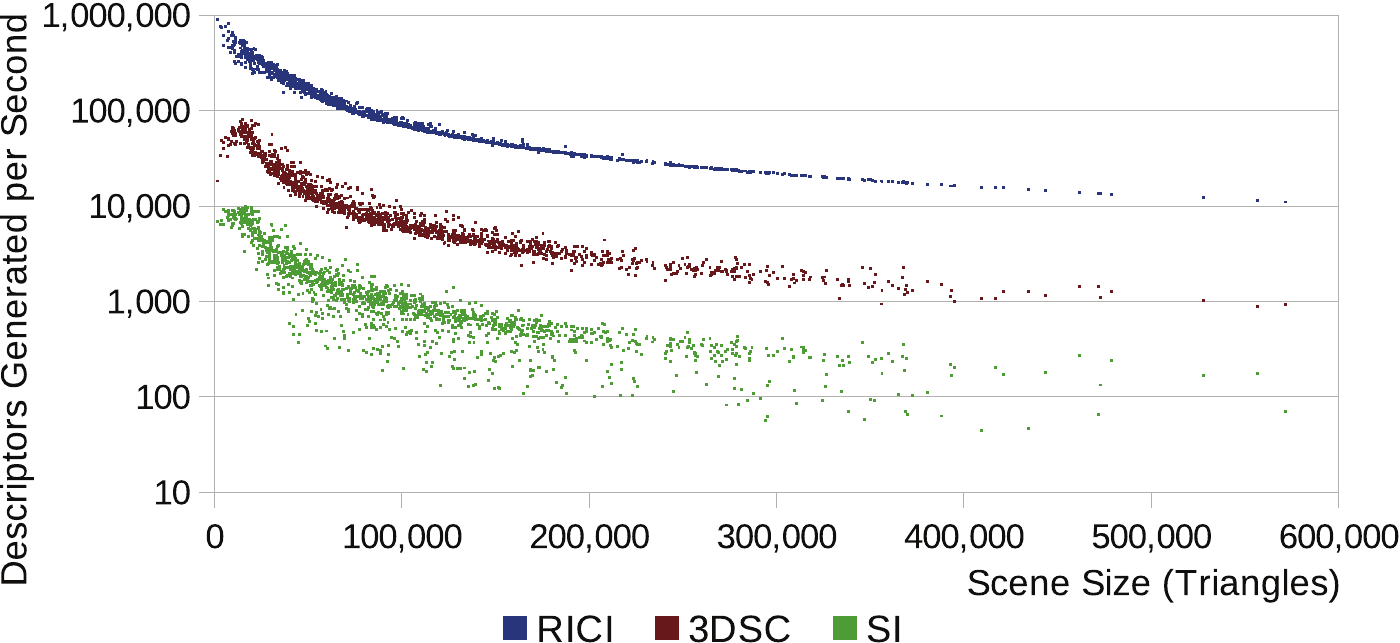}
    \caption{\small Relationship between the number of triangles present in the scene, and the rate at which our implementations generate RICI, SI, and 3DSC descriptors.}
    \label{fig:results_generation_rate}
\end{figure}

Figure \ref{fig:results_generation_rate} shows the difference in the rate at which the RICI, SI, and 3DSC descriptors are generated. As can be seen, the RICI is approximately one order of magnitude faster than the 3DSC, and two orders faster than the SI for the given settings.

\subsubsection{Performance of Point Projection Algorithm}

The largest portion of the computational effort involved in the RICI and SI generation algorithms require projecting points into cylindrical coordinate space. We have proposed an efficient algorithm for this, as outlined in Section \ref{subsubsec:projectionalgorithm}.

A similar algorithm is included in Point Cloud Library \cite{Rusu_ICRA2011_PCL}, as part of the Spin Image generation implementation. To the best of our knowledge, this was up to now the most efficient implementation available. We therefore compare our projection algorithm against this previous work. 

We evaluate both algorithms using a microbenchmark which projects a sequence of $1 \cdot 10^{9}$ randomly generated points. To ensure a fair comparison, all code unrelated to point projection has been removed from the Point Cloud Library SI generation implementation. The results are shown in Table \ref{table:execution_time_projection}.

It's worth noting that points are projected into cylindrical coordinates relative to the same oriented point. Our method can therefore precompute the values of $N_{ax}$, $N_{ay}$, $N_{bx}$, and $N_{bz}$, as outlined in section \ref{subsubsec:projectionalgorithm}. Both methods were tested on an Intel Core i7-8750H CPU.  

\begin{table}
\centering
\begin{tabular}{|l|l|}
\hline
PCL ($s$) & Proposed method ($s$) \\ \hline
7.559   & 3.084          \\ \hline       
\end{tabular}
\caption{Point projection algorithm average execution times for projecting $1\cdot10^{9}$ points.}
\label{table:execution_time_projection}
\end{table}

\subsection{Matching Rate}

\begin{figure}
    \centering
    \includegraphics[width=8cm]{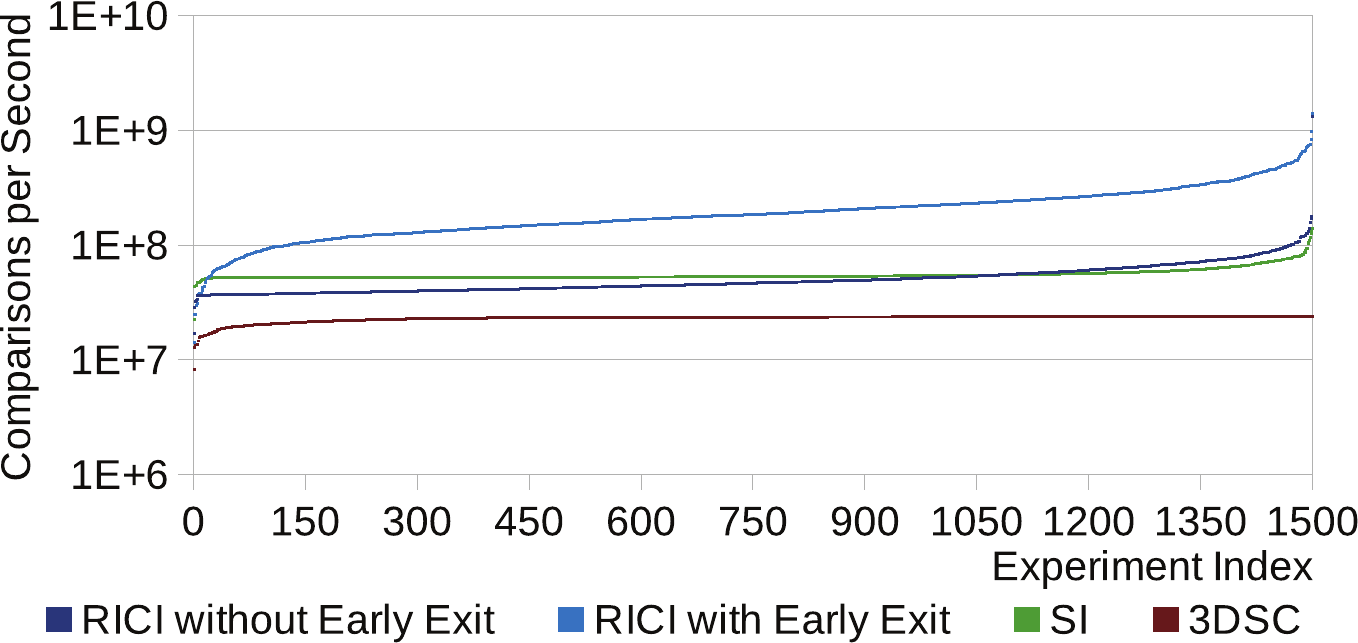}
    \caption{\small Image matching rates in a scene with 5 objects. For clarity, each sequence has been sorted individually to produce a monotonically increasing curve.}
    \label{fig:results_matching_performance}
\end{figure}

The rates of evaluating the distance functions for each method are shown in Figure \ref{fig:results_matching_performance}. As can be seen, the RICI distance function's execution times are similar to the SI's Pearson correlation coefficient, while 3DSC is significantly slower. 

For all methods, the bandwidth of the GPU memory bus is the main factor limiting the comparison rate. As our proposed distance function relies on computing the difference between neighbouring pixels, this would in a naive implementation, have required double the bandwidth. Instead, we use specialised ``shuffle instructions'' to read the value of neighbouring pixels without having to resort to another memory transaction, thereby halving the needed memory bandwidth. The result is a kernel whose memory bandwidth requirements, and consequently execution time, is similar to the Pearson Correlation Coefficient used to compare Spin Images. 

We further optimised our implementation by using an early exit condition. Since the distance score can only go up for every subsequent pixel being processed, if the only objective is determining whether the distance between two images is smaller than some given threshold distance (as is the case in many retrieval applications), it is possible to cease execution when a predetermined distance threshold is exceeded. In our clutterbox experiment, this threshold can be trivially precomputed. Utilising this early exit condition resulted on average in a 4.2 times speedup over the SI distance function. 

\section{Observations and Discussion}

There are several topics and observations that may be relevant for the interpretation of the presented results.

\subsection{Analysis of Experimental Results}

\begin{figure*}[h]
    \centering
    \begin{subfigure}[t]{0.2762\textwidth}
        \includegraphics[width=\linewidth]{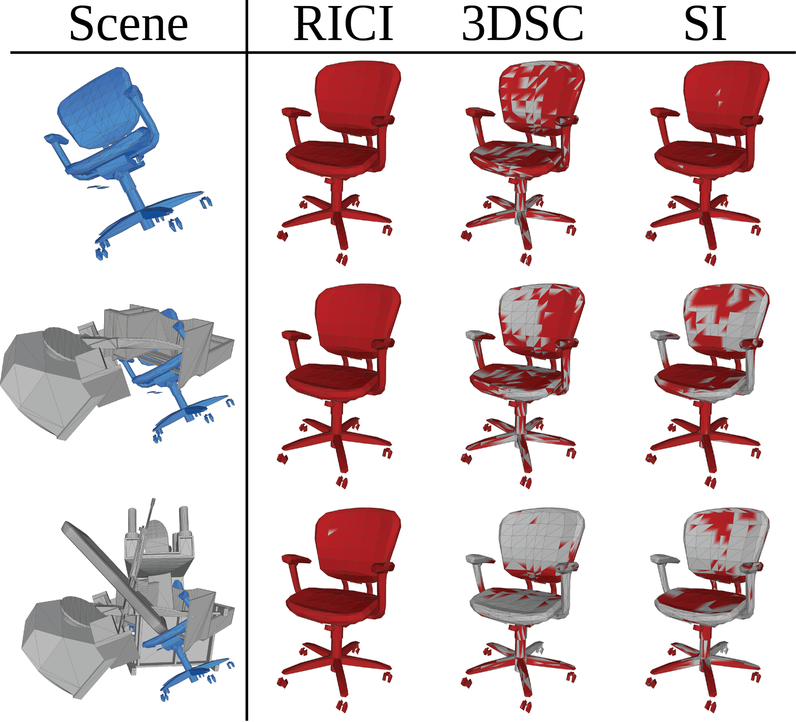}
        \caption{\label{fig:scene_lowest_drop}}
    \end{subfigure}
    \begin{subfigure}[t]{0.3146\textwidth}
        \includegraphics[width=\linewidth]{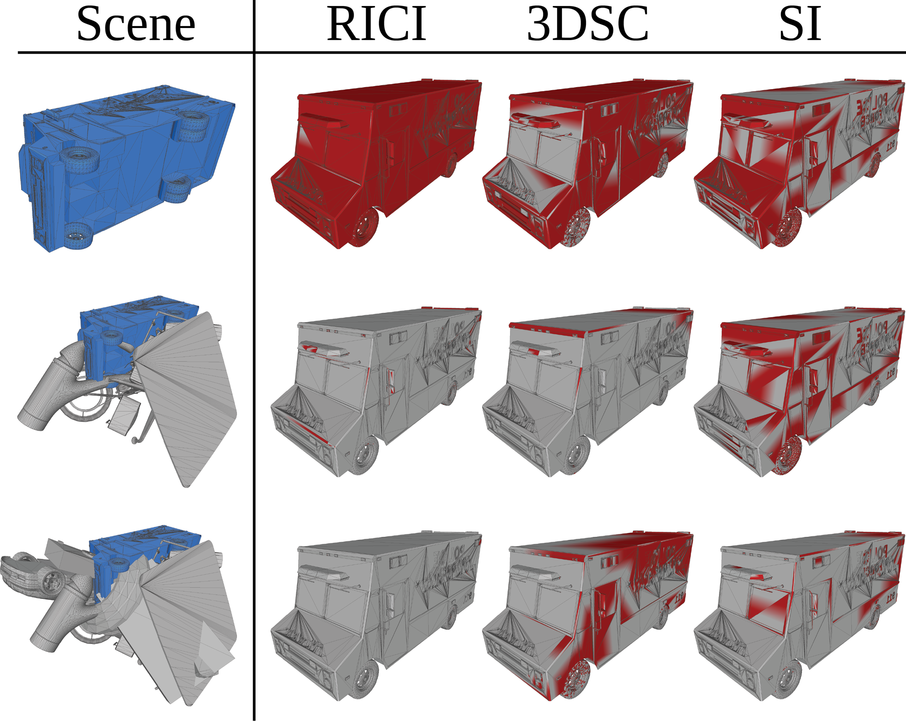}
        \caption{\label{fig:scene_largest_drop}}
    \end{subfigure}
    \begin{subfigure}[t]{0.3091\textwidth}
        \includegraphics[width=\linewidth]{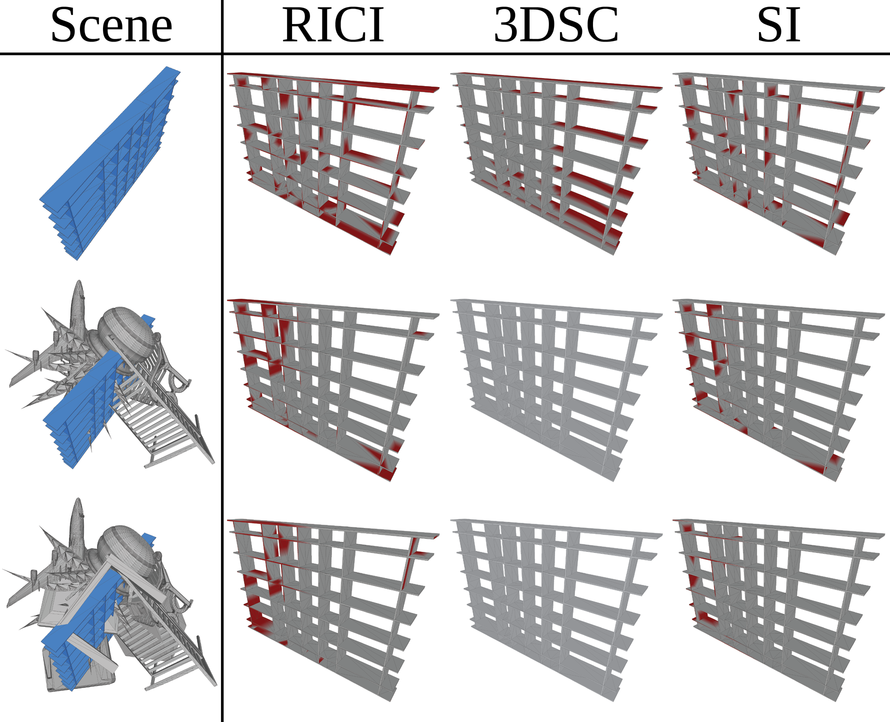}
        \caption{\label{fig:scene_worst_1}}
    \end{subfigure}
    \\ \vspace{0.3cm}
    \begin{subfigure}[t]{0.2762\textwidth}
        \includegraphics[width=\linewidth]{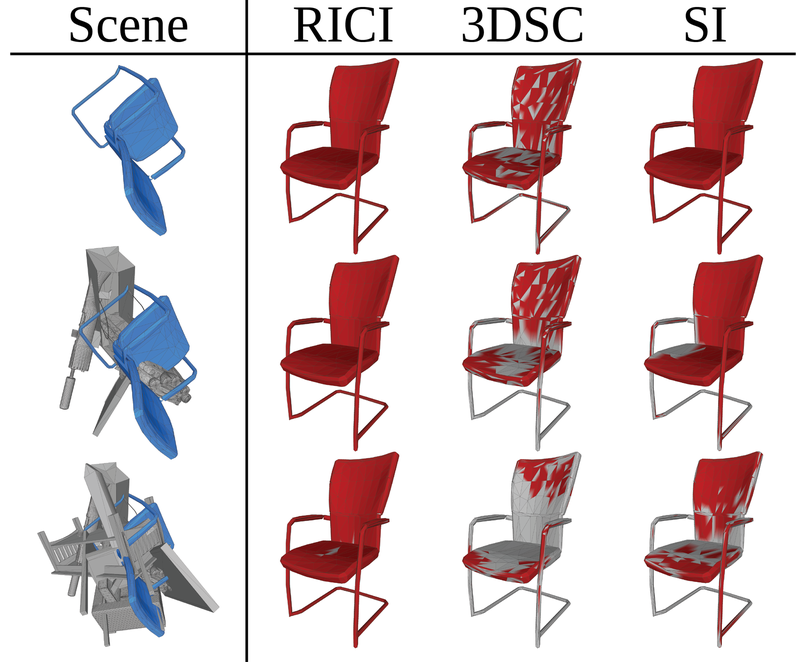}
        \caption{\label{fig:scene_best_1}}
    \end{subfigure}
    \begin{subfigure}[t]{0.3146\textwidth}
        \includegraphics[width=\linewidth]{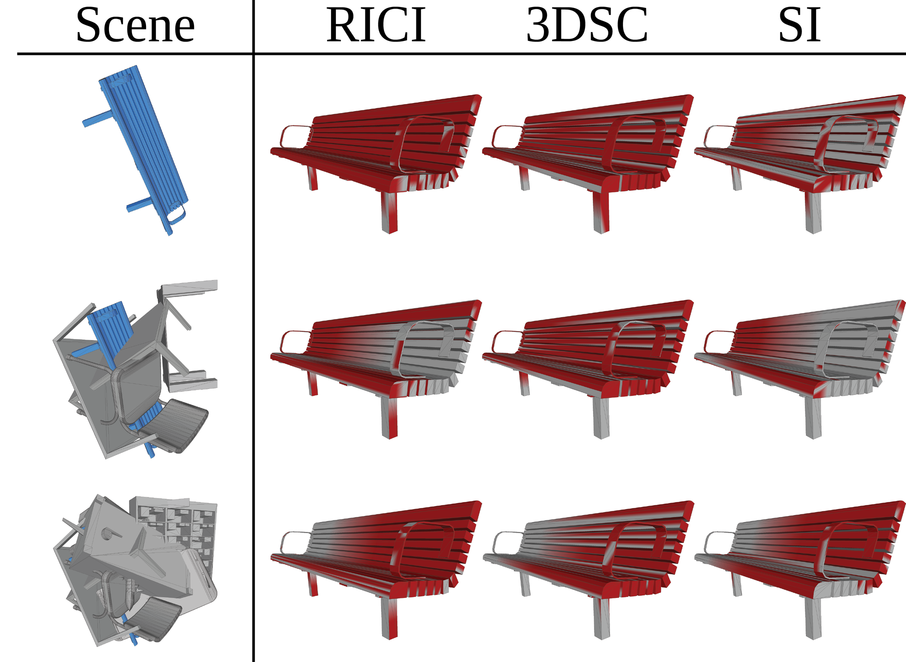}
        \caption{\label{fig:scene_perfect_average}}
    \end{subfigure}
    \begin{subfigure}[t]{0.3091\textwidth}
        \includegraphics[width=\linewidth]{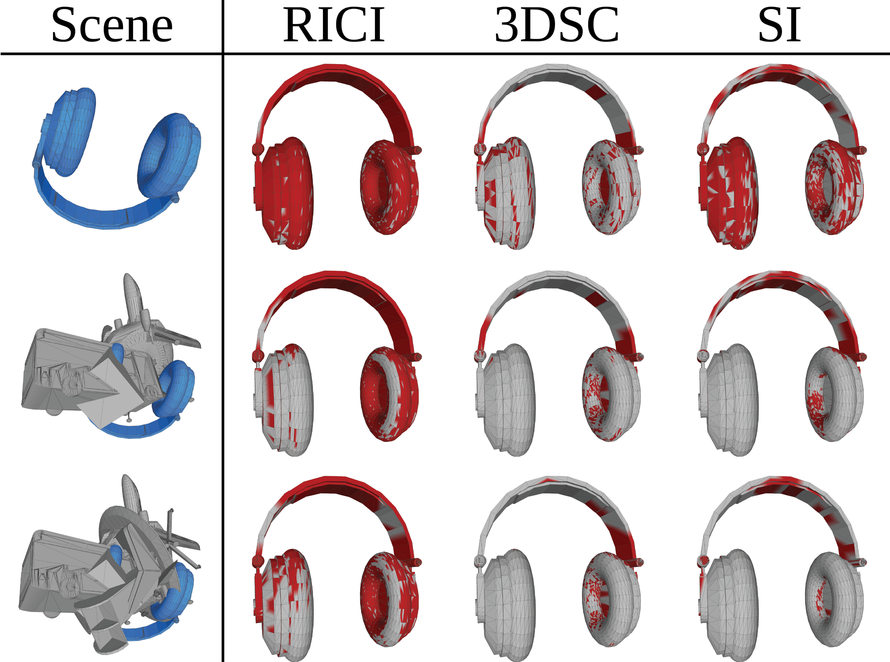}
        \caption{\label{fig:scene_clutter_better}}
    \end{subfigure}
    \caption{\small Visualised results from 6 selected experiments. For each of the 6 subfigures, the Clutterbox scene (with 1, 5, and 10 objects) is shown on the left hand side, with the reference object highlighted in blue. Vertices correctly ranked at index 0 are highlighted in red, other vertices are coloured grey.}
    \label{fig:sceneVisualisation}
\end{figure*}

While analysing the results presented in Section \ref{sec:evaluation}, we made several observations that are relevant to their interpretation. Figure \ref{fig:sceneVisualisation} contains a visualisation of a subset of these.

Figure \ref{fig:scene_lowest_drop} shows the result set where RICI experienced the smallest decrease in matching performance between 0 and 9 added clutter objects in the scene. It is also possible to observe the clutter resistant properties of RICI. The seat part of the desk chair is significantly cluttered, while the wheels experience relatively small amounts of clutter (and remain visible). All three methods are capable of reasonably recognising these exposed wheels, however, the SI and 3DSC descriptors in large part fail to recognise the cluttered seat part.

Figure \ref{fig:scene_largest_drop} shows the result set where RICI experienced the largest drop in performance between the scenes with 0 and 9 added clutter objects. The primary cause of this drop is due to the cuboid-like shape and low level of details on the police van, which causes a low number of changes in intersection counts. In turn, the produced RICI images become relatively susceptible to clutter. 

Figure \ref{fig:scene_worst_1} shows the experiment where RICI performed worst on the uncluttered reference object. The particular object, a bookshelf, has high levels of self-similarity; a property which is also, to varying degrees, present in other objects in the CAD-oriented SHREC2017 dataset. Thus any local descriptor would rank vertices belonging to self-similar regions equally and whether they end up at Rank 0 is a matter of luck. One would expect to find them within the top $s$ ranks, where $s$ is the number of self-similar vertices. On the other hand, this is a useful tool for detecting self-similar regions. 

\begin{figure*}
    \centering
    \begin{subfigure}[t]{4.0cm}
        \includegraphics[width=4.0cm]{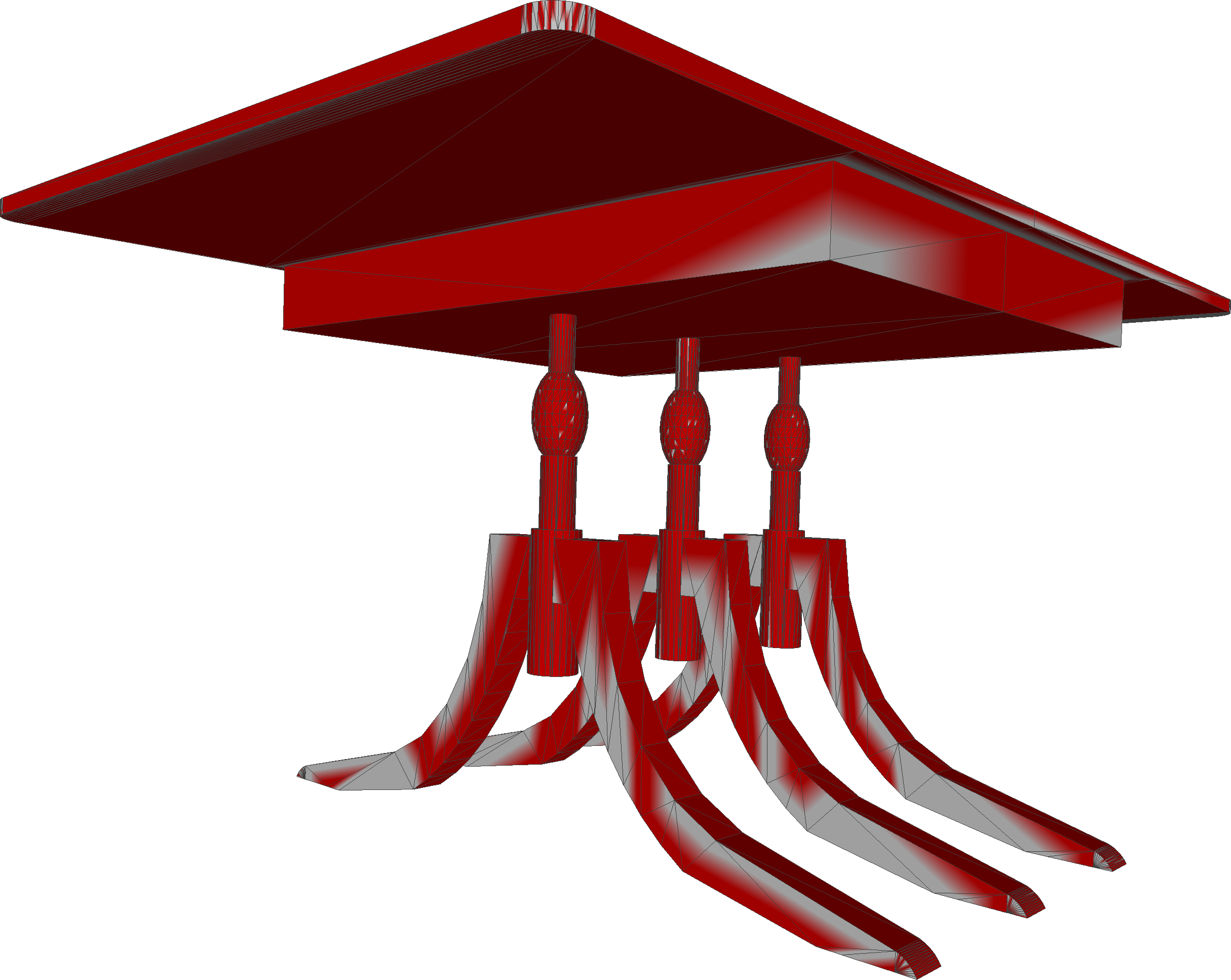}
        \caption{\label{fig:table_top1} Top rank}
    \end{subfigure} 
    %\begin{subfigure}[t]{2.0cm}
    %    \includegraphics[width=2.0cm]{images/results/table/table_top2.png}
    %    \caption{\label{fig:table_top2}}
    %\end{subfigure}
    %\begin{subfigure}[t]{2.0cm}
    %    \includegraphics[width=2.0cm]{images/results/table/table_top3.png}
    %    \caption{\label{fig:table_top3}}
    %\end{subfigure}
    \begin{subfigure}[t]{4.0cm}
        \includegraphics[width=4.0cm]{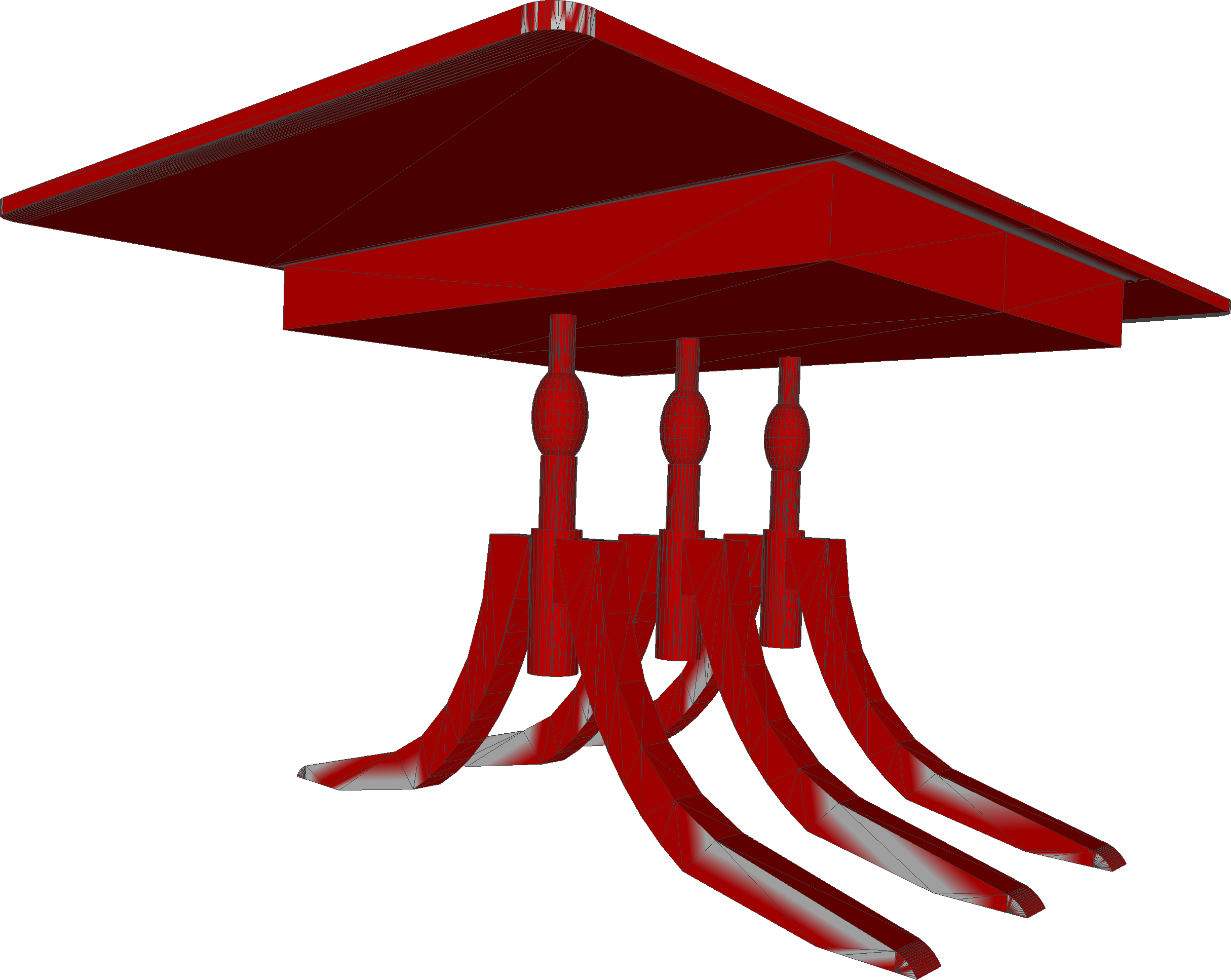}
        \caption{\label{fig:table_top4} Top 4 ranks}
    \end{subfigure}
    %\begin{subfigure}[t]{2.0cm}
    %    \includegraphics[width=2.0cm]{images/results/table/table_top5.png}
    %    \caption{\label{fig:table_top5}}
    %\end{subfigure}
    \begin{subfigure}[t]{4.0cm}
        \includegraphics[width=4.0cm]{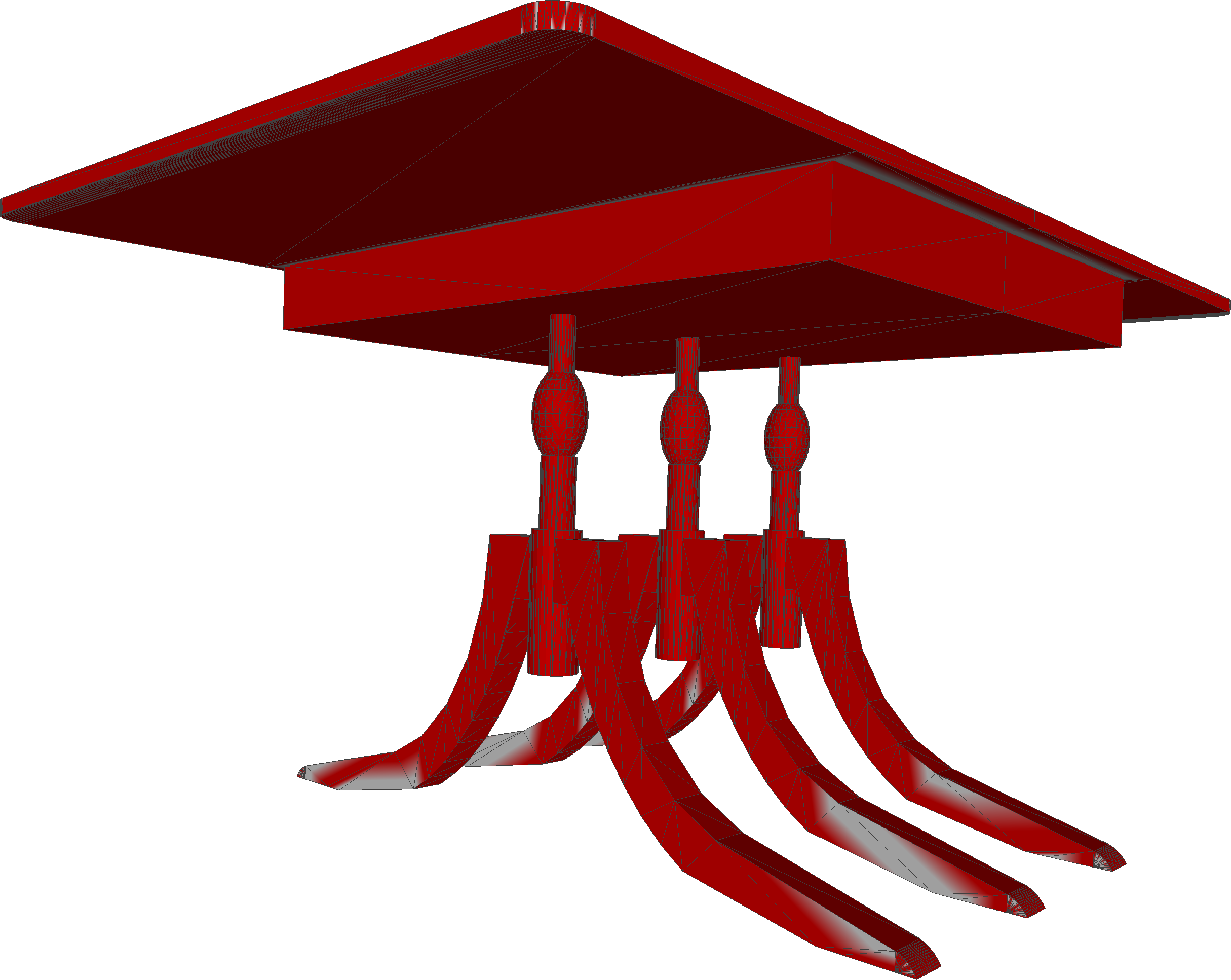}
        \caption{\label{fig:table_top6} Top 6 ranks}
    \end{subfigure}
    %\begin{subfigure}[t]{2.0cm}
    %    \includegraphics[width=2.0cm]{images/results/table/table_top7.png}
    %    \caption{\label{fig:table_top7}}
    %\end{subfigure}
    %\begin{subfigure}[t]{4.5cm}
    %    \includegraphics[width=4.5cm]{images/results/table/table_top8.png}
    %    \caption{\label{fig:table_top8} Top 8}
    %\end{subfigure}
    %\begin{subfigure}[t]{2.0cm}
    %    \includegraphics[width=2.0cm]{images/results/table/table_top9.png}
    %    \caption{\label{fig:table_top9}}
    %\end{subfigure}
    %\begin{subfigure}[t]{2.0cm}
    %    \includegraphics[width=2.0cm]{images/results/table/table_top10.png}
    %    \caption{\label{fig:table_top10}}
    %\end{subfigure}
    %\begin{subfigure}[t]{2.0cm}
    %    \includegraphics[width=2.0cm]{images/results/table/table_top11.png}
    %    \caption{\label{fig:table_top11}}
    %\end{subfigure}
    \begin{subfigure}[t]{4.0cm}
        \includegraphics[width=4.0cm]{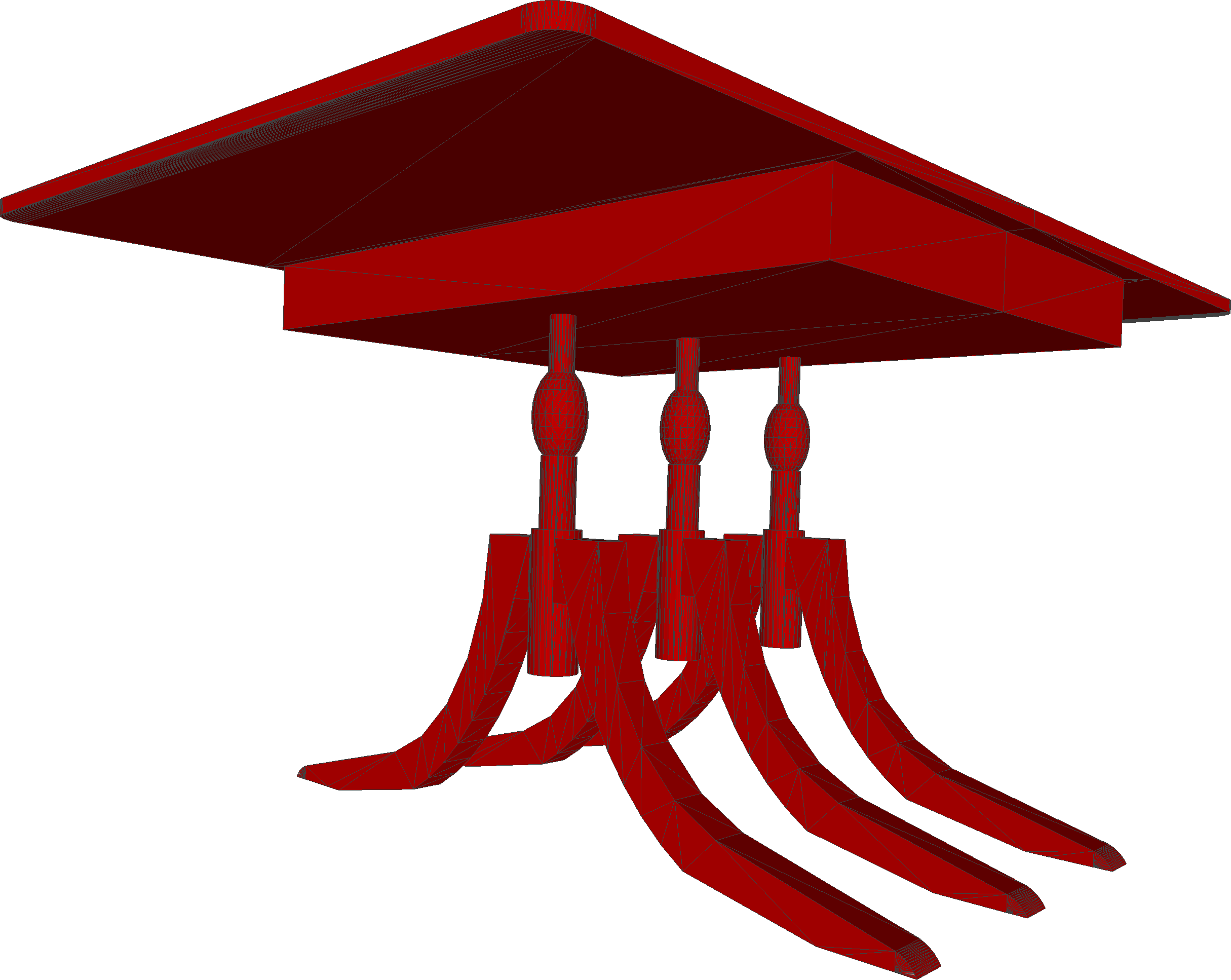}
        \caption{\label{fig:table_top12} Top 12 ranks}
    \end{subfigure}
    \caption{\small Symmetric object whose vertices were present in the top $s$ ranks of the search results, with varying values of $s$.}
    \label{fig:topNTables}
\end{figure*}

To investigate this further we visualised the results of an experiment where the reference object had countable symmetric features, as shown in Figure \ref{fig:topNTables}. As opposed to Figure \ref{fig:sceneVisualisation}, we highlighted in red those vertices that were detected in the top $s$ ranks instead of only rank 0. For instance, vertices in the table's legs are expected to constitute 12 self-similar partitions (6 legs with a symmetric front and backside each), which are all detected in the top 12 results, as shown in Figure \ref{fig:table_top12}. Also all vertices in the base of the tabletop are correctly detected within the top 4 results (4-way symmetry). 

In contrast to Figure \ref{fig:scene_worst_1}, Figure \ref{fig:scene_best_1} shows the experiment in which RICI had the highest recognition rate in the uncluttered scene. Little matching performance is lost after adding significant amounts of clutter.

In Figure \ref{fig:scene_perfect_average} the experiment whose drop in matching performance was closest to the total average of all performed 1500 experiments is shown. Worth noting here is the relatively low drop in recognition performance between the uncluttered scene, and the scene with 9 added clutter objects.

Finally, in Figure \ref{fig:scene_clutter_better} a rare phenomenon is shown where matching performance slightly improves between 4 and 9 added clutter objects.

\subsection{Performance of 3DSC}
\label{subsec:settings_discussion}

As can be seen in Figure \ref{fig:results_RICI_vs_si}, in contrast to the results obtained in previous work \cite{frome2004recognizing} \cite{guo2016comprehensive}, the SI generally outperforms the 3DSC descriptor. The primary cause of this is that in previous work, the SI resolution was set to the 15x15 bins used originally by Johnson et al. \cite{johnson1999using}. In contrast, we used a resolution of 64x64 bins for parity with the RICI descriptor, which we also consider to be a resolution more suitable to the capabilities of modern processors. This significant increase in resolution meant the SI descriptor in our testing performs better than 3DSC with our chosen settings.

The decision to use the same bin dimensions for 3DSC as in previous work was primarily motivated by a tradeoff between comparison performance and GPU hardware limitations. Our implementation makes use of shared memory when comparing 3DSC descriptors, due to the needle and haystack descriptor both being accessed once for each radial division. Current GPU shared memory pools allow fitting of approximately 2 image pairs sized at default settings simultaneously, which implies the number of bins can either be left intact, or doubled, or performance can be expected to be suboptimal. While it would be possible to double the number of bins in the 3DSC descriptor (which would make its memory requirements equal to the SI and RICI) leading to an increase in matching performance, the matching rate would decrease below acceptable levels because of the distance algorithm used. We therefore consider the used settings to be the best balance between quality and execution time for 3DSC.

\section{Conclusion}

In this paper, a clutter resistant shape descriptor, RICI, is presented and evaluated using a novel evaluation framework for such descriptors, called the clutterbox experiment. Novel algorithms for cylindrical coordinate projection, circle-triangle intersection, and the rasterization of triangles in cylindrical coordinates were presented. The largest quantitative evaluation of the SI, 3DSC, and RICI methods to date is also made, along with a useful observation for the SI support angle.  

The main advantages of RICI are its noise-free nature and generation speed, while the related distance function makes it clutter resistant. We anticipate that the proposed clutterbox experiment, which is being made public, will aid future benchmarking of shape descriptors for cluttered scenes.

\section{Acknowledgements}

The authors would like to thank the HPC-Lab leader and PI behind the "Tensor-GPU" project, Prof. Anne C. Elster, for access to the Nvidia DGX-2 system used in the experiments performed as part of this paper. Additionally, the authors would like to thank the IDUN cluster at NTNU for the provision of additional computing resources.

\printbibliography

\end{document}